\documentclass[letterpaper]{article} % DO NOT CHANGE THIS
\usepackage{aaai25}  % DO NOT CHANGE THIS
\usepackage{times}  % DO NOT CHANGE THIS
\usepackage{helvet}  % DO NOT CHANGE THIS
\usepackage{courier}  % DO NOT CHANGE THIS
\usepackage[hyphens]{url}  % DO NOT CHANGE THIS
\usepackage{graphicx} % DO NOT CHANGE THIS
\usepackage{natbib}  % DO NOT CHANGE THIS AND DO NOT ADD ANY OPTIONS TO IT
\usepackage{caption} % DO NOT CHANGE THIS AND DO NOT ADD ANY OPTIONS TO IT
\usepackage{algorithm}
\usepackage{algorithmic}
\usepackage{amsmath}
\usepackage{multirow}
\usepackage{booktabs}
\usepackage{cleveref}
\usepackage{appendix}
\usepackage{newfloat}
\usepackage{listings}
\DeclareCaptionStyle{ruled}{labelfont=normalfont,labelsep=colon,strut=off} % DO NOT CHANGE THIS
\floatstyle{ruled}
\newfloat{listing}{tb}{lst}{}
\floatname{listing}{Listing}
\title{Target-Driven Distillation: Consistency Distillation with Target Timestep Selection and Decoupled Guidance}
\author{
    Cunzheng Wang\textsuperscript{\rm 1, 2},
    Ziyuan Guo\textsuperscript{\rm 2},
    Yuxuan Duan\textsuperscript{\rm 2, 3},
    Huaxia Li\textsuperscript{\rm 2},
    Nemo Chen\textsuperscript{\rm 2},
    Xu Tang\textsuperscript{\rm 2},
    Yao Hu\textsuperscript{\rm 2}
}
\affiliations{
    \textsuperscript{\rm 1}School of Software Technology, Zhejiang University \\
    \textsuperscript{\rm 2}Xiaohongshu \\
    \textsuperscript{\rm 3} Shanghai Jiao Tong University

}

\usepackage{bibentry}
\begin{document}

\twocolumn[{
\renewcommand\twocolumn[1][]{#1}
\maketitle
\begin{center}
    \captionsetup{type=figure}
    \includegraphics[width=1.\linewidth]{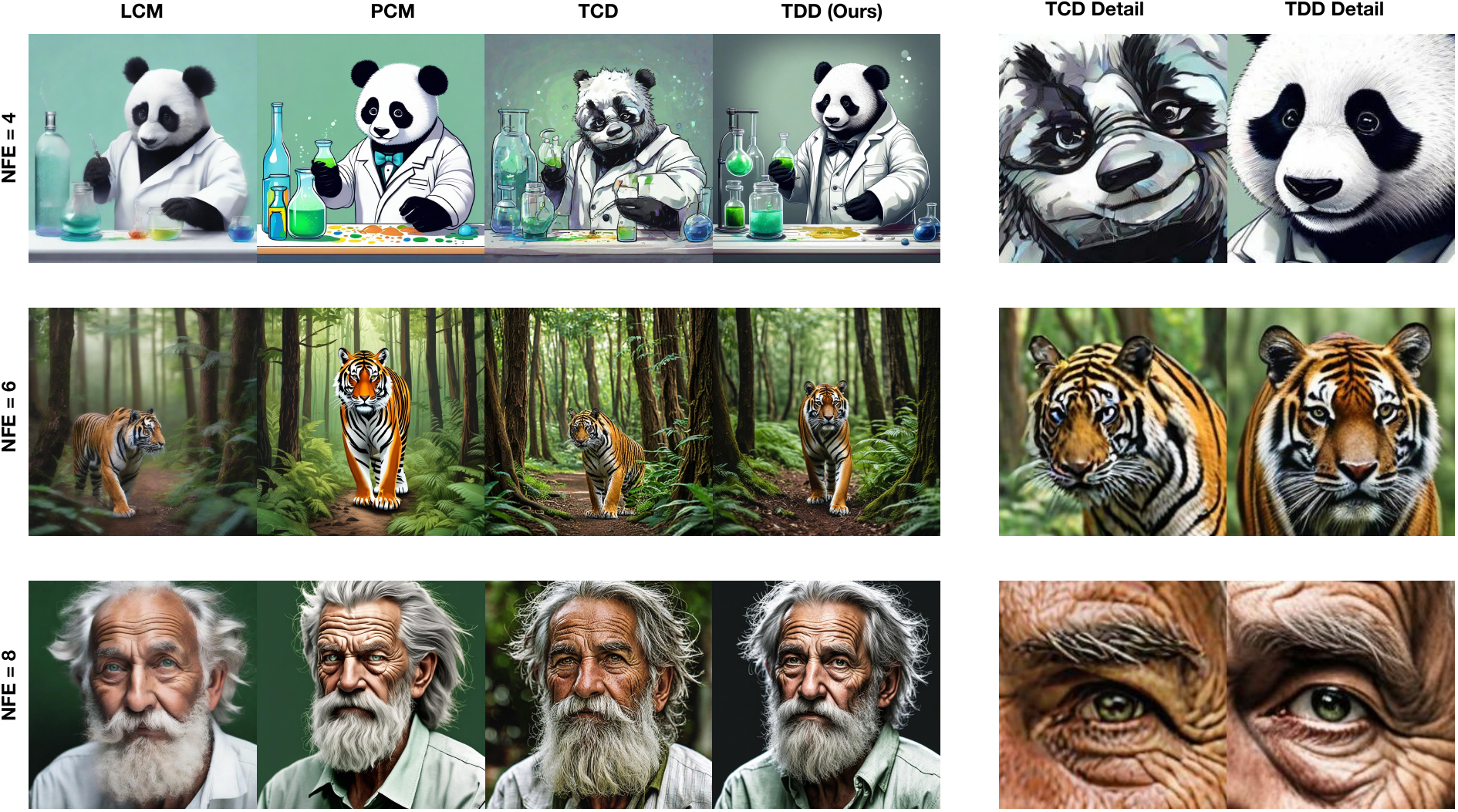}
    \caption{Visual comparison among different methods. Additionally, we have released a detailed comparison between our method and TCD. Our method demonstrates advantages in both image complexity and clarity.}
\end{center}
}]
%\maketitle

\begin{abstract}
Consistency distillation methods have demonstrated significant success in accelerating generative tasks of diffusion models. However, since previous consistency distillation methods use simple and straightforward strategies in selecting target timesteps, they usually struggle with blurs and detail losses in generated images. To address these limitations, we introduce Target-Driven Distillation (TDD), which (1) adopts a delicate selection strategy of target timesteps, increasing the training efficiency; (2) utilizes decoupled guidances during training, making TDD open to post-tuning on guidance scale during inference periods; (3) can be optionally equipped with non-equidistant sampling and $\mathbf{x}_0$ clipping, enabling a more flexible and accurate way for image sampling.
Experiments verify that TDD achieves state-of-the-art performance in few-step generation, offering a better choice among consistency distillation models.
\end{abstract}
\section{Introduction}
\begin{figure}[t]
\centering
\includegraphics[width=1.\columnwidth]{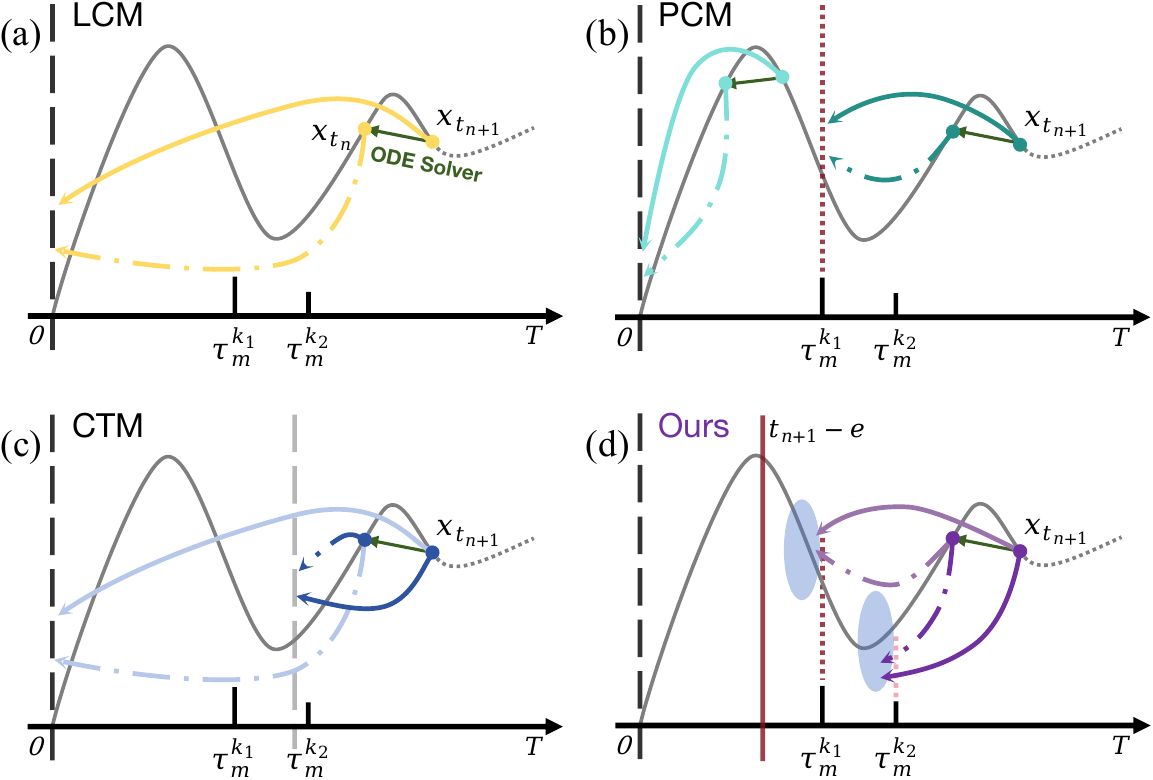} 
\caption{Comparison of different distillation methods. $ \tau^{k_1}_m $ and $ \tau^{k_2}_m $ represent a target timestep when divided into $k_1$ and $k_2$, respectively. LCM (a) and PCM (b) are examples of single-target distillation, where $ \mathbf{x}_{t_{n}} $ corresponds to only one target timestep. In contrast, CTM (c) and ours (d) are multi-target distillation methods, where $ \mathbf{x}_{t_{n}} $ can correspond to multiple target timesteps.}
\label{fig:train_diff}
\end{figure}

Diffusion models \cite{sohl2015deep, song2019generative, karras2022elucidating} have demonstrated exceptional performance in image generation, producing high-quality and diverse images. Unlike previous models like GANs \cite{goodfellow2014generative, karras2019style} or VAEs \cite{kingma2013auto, sohn2015learning}, diffusion models are good at modeling complex image distributions and conditioning on non-label conditions such as free-form text prompts.
However, since diffusion models adopt iterative denoising processes, they usually take substantial time when generating images. To address such challenge, consistency distillation methods \cite{song2023consistency, luo2023latent, luo2023lcm, kim2023consistency, zheng2024trajectory, wang2024phased} have been proposed as effective strategies to accelerate generation while maintaining image quality.
These methods distill pretrained diffusion models following the \textit{self-consistency property} \textit{i.e.} the predicted results from any two neighboring timesteps towards the same target timestep are regularized to be the same. According to the choices of target timesteps, recent consistency distillation methods can be categorized as \textit{single-target distillation} and \textit{multi-target distillation}, illustrated in \Cref{fig:train_diff}.

Single-target distillation methods follow a one-to-one mapping when choosing target timesteps, that is, they always choose \textit{the same} target timestep each time they come to a certain timestep along the trajectory of PF-ODE \cite{song2020score}. One straightforward choice is mapping any timestep to the final timestep at 0 \cite{song2023consistency, luo2023latent}. However, these methods usually suffer from the accumulated error of long-distance predictions. Another choice is evenly partitioning the full trajectory into several sub-trajectories and mapping a timestep to the end of the sub-trajectory it belongs to \cite{wang2024phased}. Although the error can be reduced by shortening the predicting distances when training, the image quality will be suboptimal when adopting a schedule with a different number of sub-trajectories during inference periods.

On the other hand, multi-target distillation methods follow a one-to-multiple mapping, that is, \textit{possibly different} target timesteps may be chosen each time they come to a certain timestep. A typical choice is mapping the current timestep to a random target timestep ahead \cite{kim2023consistency, zheng2024trajectory}. Theoretically, these methods are trained to predict from any to any timestep, thus may generally achieve good performance under different schedules. Yet, practically most of these predictions are redundant since we will never go through them under common denoising schedules. Hence, multi-target distillation methods usually require a high time budget to train.

To mitigate the aforementioned issues, we propose \textbf{Target-Driven Distillation (TDD)}, a multi-target approach that emphasizes delicately selected target timesteps during distillation processes. Our method involves three key designs: 
\textbf{Firstly,} for any timestep, it selects a nearby timestep forward that falls into a few-step equidistant denoising schedule of a predefined set of schedules (\textit{e.g.} 4--8 steps), which eliminates long-distance predictions while only focusing on the timesteps we will probably pass through during inference periods under different schedules. Also, TDD incorporates a stochastic offset that further pushes the selected timestep ahead towards the final target timestep, in order to accommodate non-deterministic sampling such as \(\gamma\)-sampling \cite{kim2023consistency}.
\textbf{Secondly,} while distilling classifier-free guidance (CFG) \cite{ho2022classifier} into the distilled models, to align with the standard training process using CFG, TDD additionally replaces a portion of the text conditions with unconditional (\textit{i.e.} empty) prompts. With such a design, TDD is open to a proposed inference-time tuning technique on guidance scale, allowing user-specified balances between the accuracy and the richness of image contents conditioned on text prompts.
\textbf{Finally,} TDD is optionally equipped with a non-equidistant sampling method doing short-distance predictions at initial steps and long-distance ones at later steps, which helps to improve overall image quality. Additionally, TDD adopts $\mathbf{x}_0$ clipping to prevent out-of-bound predictions and address the overexposure issue.

Our contributions are summarized as follows:
\begin{itemize}
    \item We provide a taxonomy on consistency distillation models, classifying previous works as \textit{single-target} and \textit{multi-target} distillation methods.
    \item We propose Target-Driven Distillation, which highlights target timestep selection and decoupled guidance during distillation processes.
    \item We present extensive experiments to validate the effectiveness of our proposed distillation method.
\end{itemize}

\section{Related Work}
Diffusion models\cite{sohl2015deep, song2019generative, karras2022elucidating} have demonstrated significant advantages in high-quality image synthesis \cite{ramesh2022hierarchical, rombach2022high, dhariwal2021diffusion}, image editing \cite{meng2021sdedit, saharia2022palette, balaji2022ediff}, and specialized tasks such as layout generation \cite{zheng2023layoutdiffusion, wu2024spherediffusion}. However, their multi-step iterative process incurs significant computational costs, hindering real-time applications. Beyond developing faster samplers \cite{song2020denoising,lu2022dpm, lu2022dpm++, zhang2022fast}, there is growing interest in model distillation approaches \cite{sauer2023adversarial, liu2022flow, sauer2024fast, yin2024one}. Among these, distillation methods based on consistency models have proven particularly effective in accelerating processes while preserving output similarity between the original and distilled models.

\citeauthor{song2023consistency} introduced the concept of consistency models, which emphasize the importance of achieving self-consistency across arbitrary pairs of points on the same probability flow ordinary differential equation (PF-ODE) trajectory \cite{song2020score}. This approach is particularly effective when distilled from a teacher model or when incorporating modules like LCM-LoRA \cite{luo2023lcm}, which can achieve few-step generation with minimal retraining resources.

However, a key limitation of these models is the increased learning difficulty when mapping points further from timestep 0, leading to suboptimal performance when mapping from pure noise in a single step. Phased Consistency Models (PCM) \cite{wang2024phased} address this by dividing the ODE trajectory into multiple sub-trajectories, reducing learning difficulty by mapping each point within a sub-trajectory to its initial point. However, in these methods, each point is mapped to a unique target timestep during distillation, resulting in suboptimal inference when using other timesteps.

Recent advancements, such as Consistency Trajectory Models (CTM) \cite{kim2023consistency} and Trajectory Consistency Distillation (TCD) \cite{zheng2024trajectory}, aim to overcome this by enabling consistency models to perform anytime-to-anytime jumps, allowing all points between timestep 0 and the inference timestep to be used as target timesteps. However, the inclusion of numerous unused target timesteps reduces training efficiency and makes the model less sensitive to fewer-step denoising timesteps.

\section{Method}

In this section, we will first deliver some preliminaries in \cref{sec:pre}, followed by detailed descriptions of our proposed Target-Driven Distillation in \cref{sec:select,sec:cfg,sec:sample}.

\subsection{Preliminaries}
\label{sec:pre}

\subsubsection{Diffusion Model}
Diffusion models constitute a category of generative models that draw inspiration from thermodynamics and stochastic processes, encompass both a forward process and a reverse process.
The forward process is modeled as a stochastic differential equation (SDE) \cite{song2020score, karras2022elucidating}. Let \( p_{\text{data}}(\mathbf{x}) \) denotes the data distribution and \( p_t(\mathbf{x}) \) the distribution of \(\mathbf{x}\) at time $t$. For a given set \(\{\mathbf{x}_t | t \in [0, T]\}\), the stochastic trajectory is described by:
\begin{equation}
\mathrm{d}\mathbf{x}_t = f(\mathbf{x}_t, t) \, \mathrm{d}t + g(t) \, \mathrm{d}\mathbf{w}_t,
\label{eq:1}
\end{equation}
where \( \mathbf{w}_t \) represents standard Brownian motion, \( f(\mathbf{x}_t, t) \) is the drift coefficient for deterministic changes, and \( g(t) \) is the diffusion coefficient for stochastic variations. At \( t = 0 \), we have \( p_0(\mathbf{x}) \equiv p_{\text{data}}(\mathbf{x}) \).

Any diffusion process described by an SDE can be represented by a deterministic process described by an ODE that shares identical marginal distributions, referred to as a Probability Flow ODE (PF-ODE). The PF-ODE is formulated as:
\begin{equation}
\mathrm{d}\mathbf{x} = \left[ f(\mathbf{x}, t) - \frac{1}{2}g(t)^2 \nabla_{\mathbf{x}} \log p_t(\mathbf{x}) \right] \mathrm{d}t,
\label{eq:pfode}
\end{equation}
where $ \nabla_{\mathbf{x}} \log p_t(\mathbf{x}) $ represents the gradient of the log-density of the data distribution $p_t(\mathbf{x})$, known as the score function. 
Empirically, we approximate this score function with a score model $ s_\phi(\mathbf{x}, t)$ trained via score matching techniques. Although there are numerous methods \cite{song2020denoising, lu2022dpm, lu2022dpm++, karras2022elucidating} available to solve ODE trajectories, they still necessitate a large number of sampling steps to attain high-quality generation results.

\subsubsection{Consistency Distillation}

To render a unified representation across all consistency distillation methods, we define the teacher model as $\phi$, the consistency function with the student model as $ \boldsymbol{f}_{\theta} $, the conditional prompt as $ c $, and the ODE solver $ \Phi(\cdots; \phi) $ predicting from a certain timestep $t_{n+1}$ to its previous timestep $t_n$ following an equidistant schedule from $T$ to $0$.
With a certain point $\mathbf{x}_{t_{n+1}}$ on the trajectory at timestep $t_{n+1}$, and its previous point $\hat{\mathbf{x}}^{\phi}_{t_n}$ predicted by $\Phi(\cdots; \phi)$, the core consistency loss can be formulated as
\begin{equation}
\mathcal{L}_{\text{CMs}}:= 
    \left\| \boldsymbol{f}_{\theta}(\mathbf{x}_{t_{n+1}}, t_{n+1}, \tau) - \boldsymbol{f}_{\theta^{-}}(\hat{\mathbf{x}}_{t_n}^{\phi}, t_n, \tau) \right\|_2^2,
\end{equation}
where $ \boldsymbol{f}_{\theta^{-}} $ is the consistency function with a target model updated with the exponential moving average (EMA) from the student model, and $\tau$ refers to the \textit{target timestep}. 

Among the mainstream distillation methods, the choices of $\tau$ are the most critical differences (see \Cref{fig:train_diff}). Single-target distillation methods select the same $\tau$ each time when predicting from a certain $t_{n+1}$. For example, CM \cite{song2023consistency} sets $\tau = 0$ for any timestep $t_{n+1}$, while PCM \cite{wang2024phased} segments the full trajectory into $\mathcal{K}$ (\textit{e.g.} 4) phased sub-trajectories, and chooses the next ending point:
\begin{equation}
    \tau = \max \left\{\tau \in \{{0, {\frac{T}{\mathcal{K}}}, {\frac{2T}{\mathcal{K}}}, \ldots, {\frac{(\mathcal{K}-1)T}{\mathcal{K}}} }\} \mid \tau < t_n\right\}.
\end{equation}
On the other hand, multi-target distillation methods may select different values for $\tau$ each time predicting from $t_{n+1}$. For instance, CTM \cite{kim2023consistency} selects a random $\tau$ within the interval $[0, t_n]$.

Our TDD, different from previous approaches that rely on simple trajectory segmentation or selecting from all possible $\tau$, employs a strategic selection of $\tau$, detailed in \cref{sec:select}. According to the taxonomy we provide in this work, TDD is a multi-target distillation method, yet we strive to reduce training on redundant predictions that are unnecessary for inference. 

\subsection{Target Timestep Selection}
\label{sec:select}

First, TDD pre-determines a set of equidistant denoising schedules, whose numbers of denoising steps range from $\mathcal{K}_{\min}$ to $\mathcal{K}_{\max}$, that we may adopt during inference periods. In the full trajectory of a PF-ODE from $T$ to $0$, for each $k \in [\mathcal{K}_{\min}, \mathcal{K}_{\max}]$, the corresponding schedule include timesteps $\{\tau_m^k\}_{m = 0}^{k - 1}$ where $\tau_m^k = \frac{mT}{k}$. Then, we can define the union of all the timesteps of these schedules as 
\begin{equation}
\label{eq:T}
    \mathcal{T} = \bigcup_{k=\mathcal{K}_{\min}}^{\mathcal{K}_{\max}} \{\tau_m^k\}_{m = 0}^{k - 1},
\end{equation}
which includes all the possible timesteps that we may choose as target timesteps. Note that \Cref{eq:T} is a generalized formulation, where $\mathcal{K}_{\min}=\mathcal{K}_{\max}=1$ for CM, $1 < \mathcal{K}_{\min}=\mathcal{K}_{\max} < N$ for PCM, and $\mathcal{K}_{\min}=\mathcal{K}_{\max}=N$ for CTM where $N$ is the total number of predictions within the equidistant schedule used by the ODE solver $\Phi$. As for our TDD, we cover commonly used few-step denoising schedules. For instance, typical values for $\mathcal{K}_{\min}$ and $\mathcal{K}_{\max}$ are respectively $4$ and $8$.

Based on the condition, we establish the consistency function as:
\begin{equation}
 \boldsymbol{f}: (\mathbf{x}_t, t, \tau) \mapsto \mathbf{x}_{\tau},
 \label{eq:tddfc}
\end{equation}
where \( t \in [0, T] \) and $ \tau \in \mathcal{T}$, and we expect that the predicted results to the specific target timestep $\tau$ will be consistent.

\subsubsection{Training}
Although $\mathcal{T}$ is already a selected set of timesteps, predicting to an arbitrary timestep in $\mathcal{T}$ still introduces redundancy, as it is unnecessary for the model to learn long-distance predictions from a large timestep $t$ to a small $\tau$ in the context of few-step sampling. Therefore, we introduce an additional constraint $ e = \frac{T}{\mathcal{K}_{\min}} $. This constraint further narrows possible choices at timestep $t$, reducing the learning difficulty.
Formally, we uniformly select
\begin{equation}
    \tau_m \sim \mathcal{U}(\{\tau \in \mathcal{T} | t - e \leq \tau \leq t\}).
\end{equation}
Besides, $\gamma$-sampling  \cite{kim2023consistency} is commonly used in few-step generation to introduce randomness and stabilize outputs. To accommodate this, we introduce an additional hyperparameter $\eta \in [0, 1]$. The final consistency target timesteps are selected following 
\begin{equation}
    \tilde{\tau}_{m} \sim \mathcal{U}([{(1-\eta){\tau}_{m}}, {\tau}_{m}]).
\end{equation}

Define the solution using the master teacher model $T_\phi$ from $ \mathbf{x}_{t_{n+1}} $ to $ \mathbf{x}_{t_{n}} $ with PF ODE solver as follows:
\begin{equation}
\hat{\mathbf{x}}^{\phi}_{t_n} = \Phi(\mathbf{x}_{t_{n+1}}, t_{n+1}, t_n; T_\phi),
\label{eq:xtq}
\end{equation}
where \( \Phi(\cdot \cdot \cdot; T_\phi) \) is update function and $ \hat{\mathbf{x}}^{\phi}_{t_n} $ is an accurate estimate of $ \mathbf{x}_{t_{n}} $ from $ \mathbf{x}_{t_{n+1}} $.
The loss function of TDD can be defined as:
\begin{equation}
\begin{aligned}
\mathcal{L}_{\text{TDD}}(\theta, \theta^{-}; \phi) := E[\sigma(t_n, \tilde{\tau}_{m})
    &\left\| \boldsymbol{f}_\theta(\mathbf{x}_{t_{n+1}}, t_{n+1}, \tilde{\tau}_{m}) \vphantom{\boldsymbol{f}_{\theta^{-}}(\hat{\mathbf{x}}_{t_n}^{\phi}, t_n, \tilde{\tau}_{m})} \right. \\
    &\left. - \boldsymbol{f}_{\theta^{-}}(\hat{\mathbf{x}}_{t_n}^{\phi}, t_n, \tilde{\tau}_{m}) \right\|_2^2],
\end{aligned}
\end{equation}
where the expectation is over $\mathbf{x} \sim p_{\text{data}}$, $n \sim {\mathcal{U}}[1, N-1]$, $\mathbf{x}_{t_{n+1}} \sim \mathcal{N}(\mathbf{x}; t_{n+1}^2 I)$ and $\hat{\mathbf{x}}_{t_n}^{\phi}$ is defined by \Cref{eq:xtq}. Namely, ${\mathcal{U}}[1, N-1]$ denotes a uniform distribution over 1 to $N-1$, where $N$ is a positive integer. $\sigma(\cdot, \cdot)$ is a positive weighting function, following CM, we set \( \sigma(t_n, \tilde{\tau}_{m}) \equiv 1 \). For a detailed description of our algorithm, please refer to \Cref{alg:gtcd}.

\begin{algorithm}[tb]
\caption{Target-Driven Distillation}
\label{alg:gtcd}
\textbf{Input}: dataset $\mathcal{D}$, , learning rate $\delta$, the update function of ODE solver \( \Phi(\cdot \cdot \cdot; \cdot) \), EMA rate $\mu$, noise schedule $\alpha_t, \sigma_t$, number of ODE steps $N$, fixed gudiance scale $\omega'$, empty prompt ratio $\rho$.\\
\textbf{Parameter}: initial model parameter $\theta$\\
\textbf{Output}:
\begin{algorithmic}[1] %[1] enables line numbers
\STATE \(\mathcal{T} \gets \emptyset\)
\FOR{$k \in \{\mathcal{K}_{\min}, \mathcal{K}_{\min}+1, \ldots, \mathcal{K}_{\max}\}$}
\STATE Set time steps $\tau^k_m \in \{ \tau^k_0, \tau^k_1, \ldots, \tau^k_{k-1} \}$
\STATE Add time steps to $\mathcal{T}$
\ENDFOR
\STATE let $e = \frac{T}{\mathcal{K}_{\min}}$
\REPEAT
\STATE Sample $(z, c) \sim \mathcal{D}, \tau_m \sim \mathcal{U}(\{\tau \in \mathcal{T} | t - e \leq \tau \leq t\})$
\STATE Sample $n \sim \mathcal{U}[1, N - 1], \tilde{\tau}_{m} \sim \mathcal{U}([{(1-\eta){\tau}_{m}}, {\tau}_{m}])$

\STATE Sample $\mathbf{x}_{t_{n+1}} \sim \mathcal{N}(\alpha_{t_{n+1}} \mathbf{x}, \sigma_{t_{n+1}}^2 \mathbf{I})$
\IF{probability $>$ $\rho$}
    \STATE $\begin{aligned}\hat{\mathbf{x}}^{\phi, w'}_{t_n} \gets &(1 + \omega') \Phi(\mathbf{x}_{t_{n+1}}, t_{n+1}, t_n, c; T_\phi) \\
& - \omega' \Phi(\mathbf{x}_{t_{n+1}}, t_{n+1}, t_n; T_\phi)\end{aligned}$
\ELSE
    \STATE $\hat{\mathbf{x}}^{\phi, w'}_{t_n} \gets \Phi(\mathbf{x}_{t_{n+1}}, t_{n+1}, t_n;T_\phi)
    $
\ENDIF
\STATE 
$
\mathcal{L}^{w'}_{\text{TDD}}:= 
    \left\| \boldsymbol{f}_{\theta}(\mathbf{x}_{t_{n+1}}, t_{n+1}, \tilde{\tau}_{m}) - \boldsymbol{f}_{\theta^{-}}(\hat{\mathbf{x}}^{\phi, w'}_{t_n}, t_n, \tilde{\tau}_{m}) \right\|_2^2
$
\STATE $\theta \gets \theta - \delta \nabla_{\theta} \mathcal{L}(\theta, \theta^-; \phi)$
\STATE $\theta^- \gets \text{sg}(\mu \theta^- + (1 - \mu) \theta)$
\UNTIL convergence
\end{algorithmic}
\end{algorithm}

\begin{figure*}[t]
\centering
\includegraphics[width=2.\columnwidth]{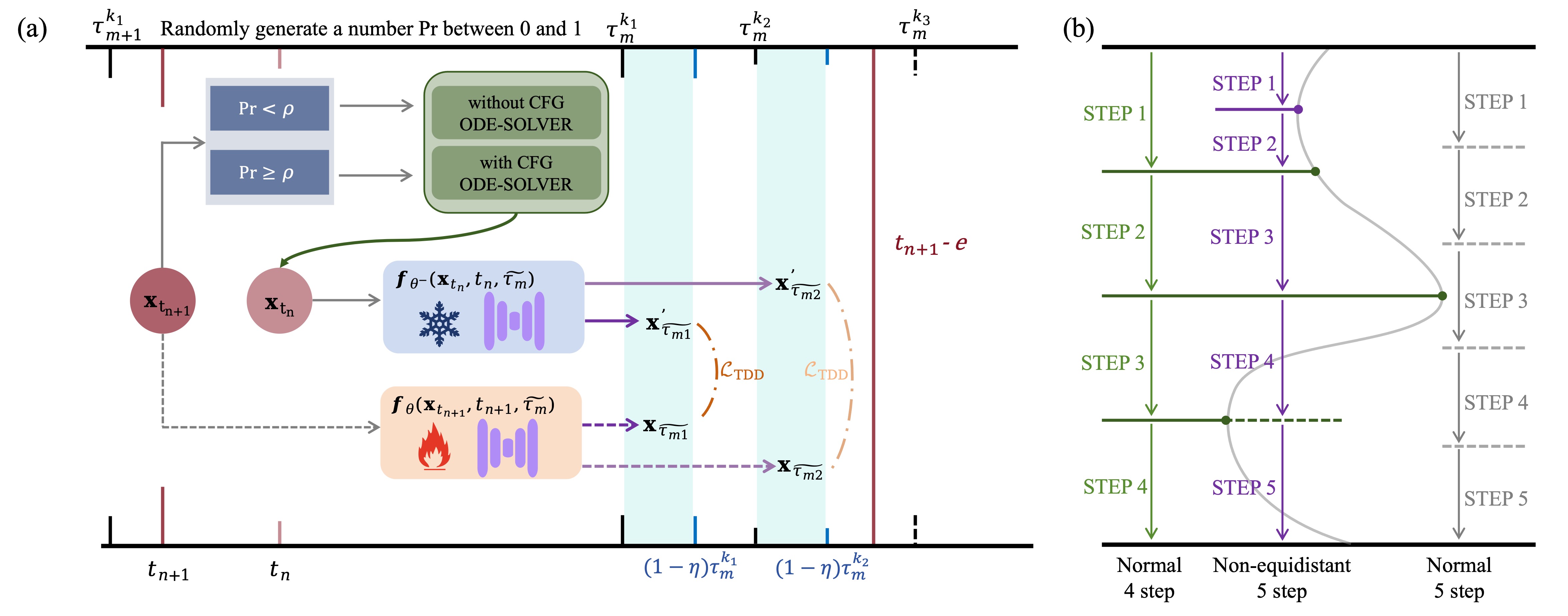} %sample_fin
\caption{Illustration of TDD distillation training and sampling processes. Fig (a) shows the distillation process, where $\tau^{k}$ represents equidistant timestep within segments. Fig (b) compares non-equidistant sampling with standard sampling for 5-step inference.}

\label{fig:sample}
\end{figure*}

\subsection{Decoupled Guidance}
\label{sec:cfg}

\subsubsection{Distillation with Decoupled Guidance}
Classifier-Free Guidance allows a model to precisely control the generation results without relying on an external classifier during the generation process, effectively modulating the influence of conditional signals. In current consistency model distillation methods, to ensure the stability of the training process, it is common to use the sample \(\hat{\mathbf{x}}^{\phi, w'}_{t_n}\) generated by the teacher model with classifier-free guidance as a reference in the optimization process for the student model's generated samples. We believe that $w'$ solely represents the diversity constraint in the distillation process, controlling the complexity and generalization of the learning target. This allows for faster learning with fewer parameters. Therefore, $w'$ should be treated separately from the CFG scale $w$ used during inference in consistency models. Therefore, regardless of whether $w' > 0$, following \cite{ho2022classifier}, it is essential to include both unconditional and conditional training samples in the training process.
Based on this, we will replace a portion of the condition with an empty prompt and not apply CFG enhancement. For conditions that are not empty, the loss function of TDD can be updated as follows:
\begin{equation}
    \mathcal{L}^{w'}_{\text{TDD}}:= 
    \left\| \boldsymbol{f}_{\theta}(\mathbf{x}_{t_{n+1}}, t_{n+1}, \tau) - \boldsymbol{f}_{\theta^{-}}(\hat{\mathbf{x}}^{\phi, w'}_{t_n}, t_n, \tau) \right\|_2^2.
\end{equation}

\begin{figure*}[t]
\centering
\includegraphics[width=2.0\columnwidth]{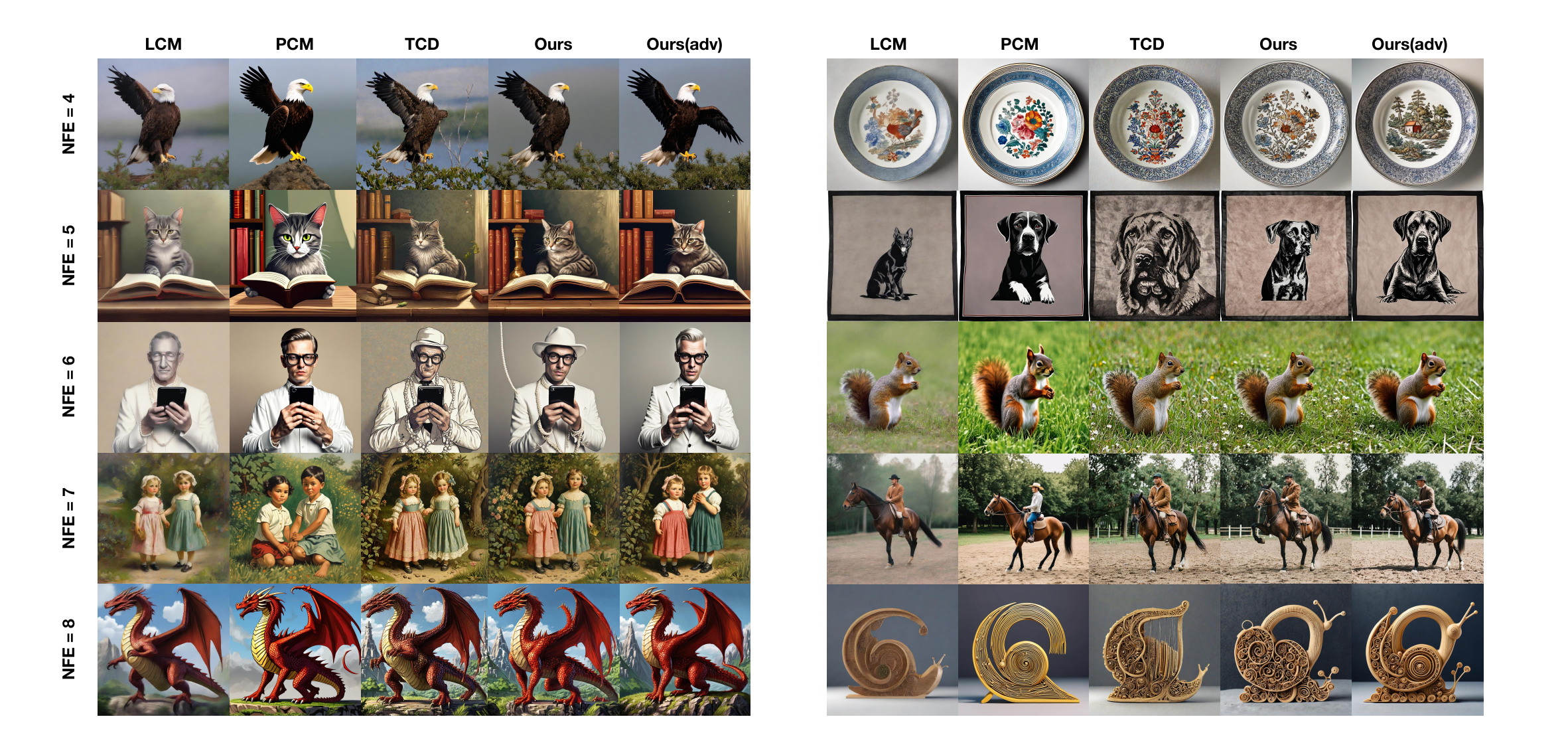}
\caption{Qualitative comparison of different methods under NFE for 4 to 8 steps.}
\label{fig:main_res}
\end{figure*}

\subsubsection{Guidance Scale Tuning}
Define $\epsilon_\theta(\mathbf{x}_t)$ as consistency model, at each inference step, the noise predicted by the model can be expressed as:
% \[
% \hat{\epsilon} = \epsilon_\theta(x_t, t) - w \sqrt{1 - \bar{\alpha}_t} \nabla_{x_t} \log p(y|x_t)
% \]
\begin{equation}
  \hat{\epsilon_w} = (1 + {w})\epsilon_\theta({\mathbf{x}}^{w'}_{t}, t, c) - {w}\epsilon_\theta({\mathbf{x}}^{w'}_{t}, t),
\end{equation}
where ${\mathbf{x}}^{w'}_{t}$ represents the input state of the consistency model at time step $t$ distilled with the diversity guidance scale $w'$.

Although setting a high value for $w'$ (e.g., $w' > 7$) can enhance certain aspects of the generated images, it simultaneously results in significantly reduced image complexity and excessively high contrast.

Is there a way to address this issue without retraining, allowing us to revert to results that enable inference with a small CFG, similar to the original model? By incorporating the unconditional into the training, we can get $
\epsilon_\theta({\mathbf{x}}^{w'}_{t}, t, c) \propto (1 + {w'})\epsilon_\phi({\mathbf{x}}_{t}, t, c) - {w'}\epsilon_\phi({\mathbf{x}}_{t}, t) 
$ and  $\epsilon_\theta({\mathbf{x}}^{w'}_{t}, t) \propto \epsilon_\phi({\mathbf{x}}_{t}, t)
$, where $\epsilon_\phi(\cdot)$ is the master model.

Denote \((1+w)\epsilon_\phi({\mathbf{x}}_{t}, t, c)-w\epsilon_\phi({\mathbf{x}}_{t}, t)\) as $\epsilon_{w}$, representing the noise inferred by the original model when using the normal guidance scale $ w $ at each time step. After simplification, we finally obtain:
\begin{equation}
    \epsilon_{w} \approx [\hat{\epsilon_w} + {w'}\epsilon_\theta({\mathbf{x}}_{t}, t)]/(1+{w'}).
\end{equation}
This equation suggests that regulating \( \hat{\epsilon_w} \) with the normal $w$ can approximate the output of the original model $ \epsilon_{w} $. For a more detailed derivation, please refer to the appendix.

In the aforementioned formula, the distillation diversity constraint $w'$ is known and fixed. The parameter $w$ can be inferred based on the standard teacher model's ratio. Furthermore, for the current consistency model, even though it has not yet learned the unguided path, this formula can still be approximately utilized for inference, as no other learning has been conducted. This can be expressed as follows:
\begin{equation}
    \epsilon'_{w} \approx [\hat{\epsilon'_w} + \overline{w'}\epsilon'_\theta({\mathbf{x}}_{t}, t)]/(1+\overline{w'}).
\end{equation}
where $\epsilon'_\theta(\cdot)$ is current consistency model and $\overline{w'}= {w'_{\min} + w'_{\max})/2}$.

\subsection{Sample}
\label{sec:sample}
Since we have trained on multiple equidistant target timesteps, we can extend this to non-equidistant sampling. By reducing the sampling interval during high-noise periods, we can mitigate the generation difficulty and achieve better synthesis results. As shown in \Cref{fig:sample} (b), we ensure that the inference process passes through the target timesteps corresponding to $ \mathcal{K}_{\min} $. As the inference steps increase, additional target timesteps are gradually inserted between these key timesteps. For example, with $ \mathcal{K}_{\min}=4 $, as the number of steps increases, we insert 8-step ($ \mathcal{K}_{\max} $) target timesteps within each adjacent 4-step target interval to enhance generation quality.

In addition, the $\gamma$-sampler proposed in CTM \cite{kim2023consistency} alternates forward and backward jumps along the solution trajectory to solve \(\mathbf{x}_0\), allowing control over the randomness ratio through $\gamma$, which can enhance generation quality to some extent. Solving for $\mathbf{x}_s$ from $\mathbf{x}_t$ can be represented as follows:
\begin{equation}
\mathbf{x}_s = \frac{\alpha_s}{\alpha_t} \mathbf{x}_t - \sigma_s \left(e^{h_s} - 1\right) \epsilon_\theta(\mathbf{x}_t, t),
\end{equation}
where $h_s=\lambda_s - \lambda_t$ and $\lambda$ represents the log signal-to-noise ratio, $\lambda=\log(\alpha/\sigma)$.However, in few-step sampling with high CFG inference, the noise distribution after the first step significantly deviates from the expected distribution. To address this, we adopt an \(\mathbf{x}_0\) formulation.To approximate $x_\theta(x_t, t) \approx x_0 = (\mathbf{x}_t - \sigma_t \epsilon)/(\alpha_t)$, we can derive:
\begin{equation}
    \mathbf{x}_s = \frac{\sigma_s}{\sigma_t} \mathbf{x}_t - \alpha_s \left(e^{-h_s} - 1\right)x_\theta(\mathbf{x}_t, t).
\end{equation}
Following prior works \cite{saharia2022photorealistic, lu2022dpm++}, we apply the clipping method $\mathcal{C}$, which clips each latent variable to the specific percentile of its absolute value and normalizes it to prevent saturation of the latent variables.
Let $\hat{\mathbf{x}_0}=\mathcal{C}(x_\theta(\mathbf{x}_t, t))$, we ultimately obtain:
\begin{equation}
    \hat{\mathbf{x}_s} = \frac{\sigma_s}{\sigma_t} \mathbf{x}_t - \alpha_s \left(e^{-h_s} - 1\right)\hat{\mathbf{x}_0}.
\end{equation}
When using $\gamma$-sampler,  the transition from timestep $t$ to  the next timestep $p$ can be expressed as
\begin{equation}
    \mathbf{x}_p = \frac{\alpha_p}{\alpha_s} \hat{\mathbf{x}_s} + \sqrt{1 - \frac{\alpha_p^2}{\alpha_s^2}} \mathbf{z}, \quad \mathbf{z} \in \mathcal{N}(0, I),
    \label{eq:x0clip}
\end{equation}
where $s=(1-\gamma)p$.

\begin{table*}[h]
\centering

\resizebox{1.\textwidth}{!}{
\begin{tabular}{l|ccccc|ccccc}
\toprule
\multirow{3}{*}{METHOD} & \multicolumn{5}{c|}{FID $\downarrow$} & \multicolumn{5}{c}{Image Complexity Score $\uparrow$} \\ \cline{2-11} 
& \multicolumn{5}{c|}{COCO-30K} & \multicolumn{5}{c}{COCO-30K} \\ \cline{2-11} 
                        & 4 steps & 5 steps & 6 steps & 7 steps & 8 steps & 4 steps & 5 steps & 6 steps & 7 steps & 8 steps  \\ \hline
LCM & 18.424 & 18.906 & 19.457 & 19.929 & 20.494 
& 0.419 & 0.417 & 0.415  & 0.412 & 0.409\\
PCM & 22.213 & 21.921 & 21.916 & 21.786 & 21.772 & 0.433 & 0.447 & 0.458  & 0.465 & 0.471 \\
TCD & \underline{17.351} & \underline{17.430} & \underline{17.535} & \underline{17.65943} & \underline{17.771}
& \textbf{0.512} & \textbf{0.521} & \textbf{0.527}  & \textbf{0.533} & \textbf{0.536} \\
Ours & \textbf{17.256} & \textbf{17.388} & \textbf{17.334} & \textbf{17.565} & \textbf{17.681} 
& \underline{0.474} & \underline{0.488} & \underline{0.499} & \underline{0.508} & \underline{0.516}\\ \bottomrule
% \midrule
\toprule
%\toprule
\multirow{3}{*}{METHOD} & \multicolumn{10}{c} {PickScore $\uparrow$} \\ \cline{2-11}
& \multicolumn{5}{c|}{PartiPrompts} & \multicolumn{5}{c}{COCO-2K} \\ \cline{2-11}
                        & 4 steps & 5 steps & 6 steps & 7 steps & 8 steps & 4 steps & 5 steps & 6 steps & 7 steps & 8 steps  \\ \hline
LCM & 22.182 & 22.226 & 22.246 & 22.251 & 22.237 & 22.143 & 22.223 & 22.268 & 22.296 & 22.297\\
PCM & 22.200 & 22.289 & 22.340 & 22.357 & 22.367
& 22.291 & 22.433 & 22.505 & 22.537 & \underline{22.543}
\\
TCD & \textbf{22.350} & 22.402 & 22.417 & 22.426 & 22.426
& 22.310 & 22.423 & 22.490 & 22.504 & 22.504
\\
Ours & \underline{22.338} & \underline{22.414} & 22.445 & 22.472 & \textbf{22.493}
& \textbf{22.396} & \underline{22.533} & \underline{22.593} & \underline{22.640} & \textbf{22.656}
\\
Ours(*) & - & \textbf{22.431} & \textbf{22.472} & \textbf{22.489} 
& - & - & \textbf{22.577} & \textbf{22.630} & \textbf{22.652} & - \\
Ours(adv) & 22.279 & 22.388 & \underline{22.450} & \underline{22.476} & \underline{22.490} & \underline{22.322} & 22.466 & 22.502 & 22.511 & 22.508 \\
\bottomrule
\end{tabular}
}
\caption{Quantitative Comparison under different metrics and datasets. Our method consistently outperforms others in FID and PickScore, achieving higher image complexity without sacrificing quality.}
\label{tab:comparison_1}
\end{table*}

\section{Experiments}

\subsection{Dateset}

We use a subset of the Laion-5B \cite{schuhmann2022laion} High-Res dataset for training. All images in the dataset have an aesthetic score above 5.5, totaling approximately 260 million. Additionally, we evaluate the performance using the COCO-2014 \cite{lin2014microsoft} validation set, split into 30k (COCO-30K) and 2k (COCO-2K) captions for assessing different metrics. We also use the PartiPrompts \cite{yu2022scaling} dataset to benchmark performance, which includes over 1600 prompts across various categories and challenge aspects.

\subsection{Backbone}
We employ SDXL \cite{podell2023sdxl} as the backbone for our experiments. Specifically, we trained a LoRA \cite{hu2021lora} through distillation.

\subsection{Metrics}
We evaluate image generation quality using Frechet Inception Distance (FID) \cite{heusel2017gans} and assess image content richness with the Image Complexity 
Score (IC) \cite{feng2022ic9600}. Additionally, we use PickScore \cite{kirstain2023pick} to measure human preference. FID and IC are tested on the COCO-30K dataset, while PickScore is evaluated on COCO-2K and PartiPrompts.

\subsection{Performance Comparison}
In this section, we present a comprehensive performance evaluation of our proposed method against several baselines, including LCM, PCM, and TCD, across the COCO-30K, PartiPrompts, and COCO-2K datasets, as detailed in \Cref{tab:comparison_1}. We introduce two methods, ``Ours" and ``Ours*", representing normal sampling and non-equidistant sampling (4-step and 8-step as normal sampling), respectively. Additionally, ``Ours(adv)" incorporates PCM's \cite{wang2024phased} adversarial process during distillation, demonstrating that our method can effectively integrate adversarial training.

As shown in \Cref{fig:main_res}, we qualitatively compare Ours and Ours(adv) with other methods across different inference steps. Our model outperforms others in image quality and text-image alignment, especially in the 4 to 8-step range. Quantitative results in show that Ours achieves the best FID, though FID values tend to increase with more steps. While our method may not always achieve the top Image Complexity (IC), it avoids generating less detailed or cluttered images, unlike CTM, as seen in \Cref{fig:main_res}. However, the high image complexity observed in CTM may also be attributed to certain visual artifacts and high-frequency noise, which we will elaborate on in the Appendix. Furthermore, evaluations using PickScore on the COCO-2K and PartiPrompt datasets show that Ours and Ours* consistently rank first or second in most cases. Overall, our method demonstrates superior performance and a well-balanced approach across metrics.

\subsection{Ablation Study}
\subsubsection{Effect of Target Timestep Selection}
\begin{figure}[t]
\centering
\includegraphics[width=1.\columnwidth]{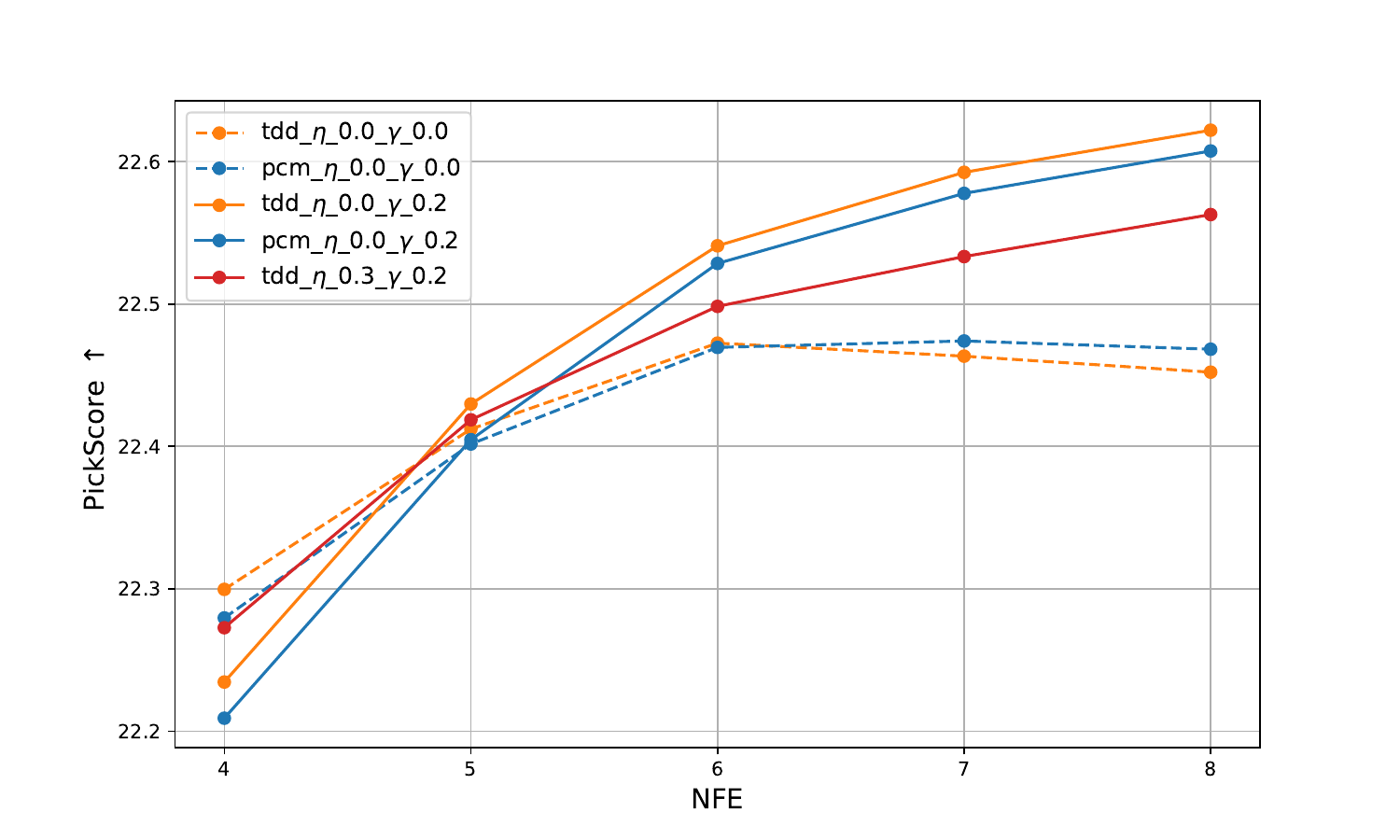}
\caption{Qualitative comparison between Target-Driven Multi-Target Distillation (TDD, 4-8 step target timesteps distillation) and Single-Target Distillation (PCM, 4-step target timesteps distillation).}
\label{ab_tdd}
\end{figure}

To demonstrate the advantages of Target-Timestep-Selection, we compared the performance of models trained on mappings required for 4-step inference (\textit{e.g.}, PCM) against those trained on mappings required for 4--8 step inference. We maintained consistent settings with a batch size of 128, a learning rate of 5e-06, and trained for a total of 15,000 steps. As shown in \Cref{ab_tdd}, when using deterministic sampling (i.e., \(\gamma=0\)), the model trained on mappings for 4--8 step inference showed only a slight advantage at 4 and 5 steps. However, when incorporating randomness into sampling (i.e., \(\gamma=0.2\)), the model trained on 4--8 step mappings outperformed the model trained on 4-step mappings across all 4--8 steps. Furthermore, when we extended the mapping range by \(\eta=0.3\) to better accommodate the randomness in sampling (as indicated by the red line in \Cref{ab_tdd}), inference with \(\gamma=0.2\) achieved a well-balanced performance across 4--8 steps, avoiding poor performance at 4--5 steps while also maintaining solid performance at 6--8 steps.
\subsubsection{Effect of Distillation with Decoupled Guidance and Guidance Scale Tuning}
\begin{figure}[t]
    \centering
    \includegraphics[width=1.\linewidth]{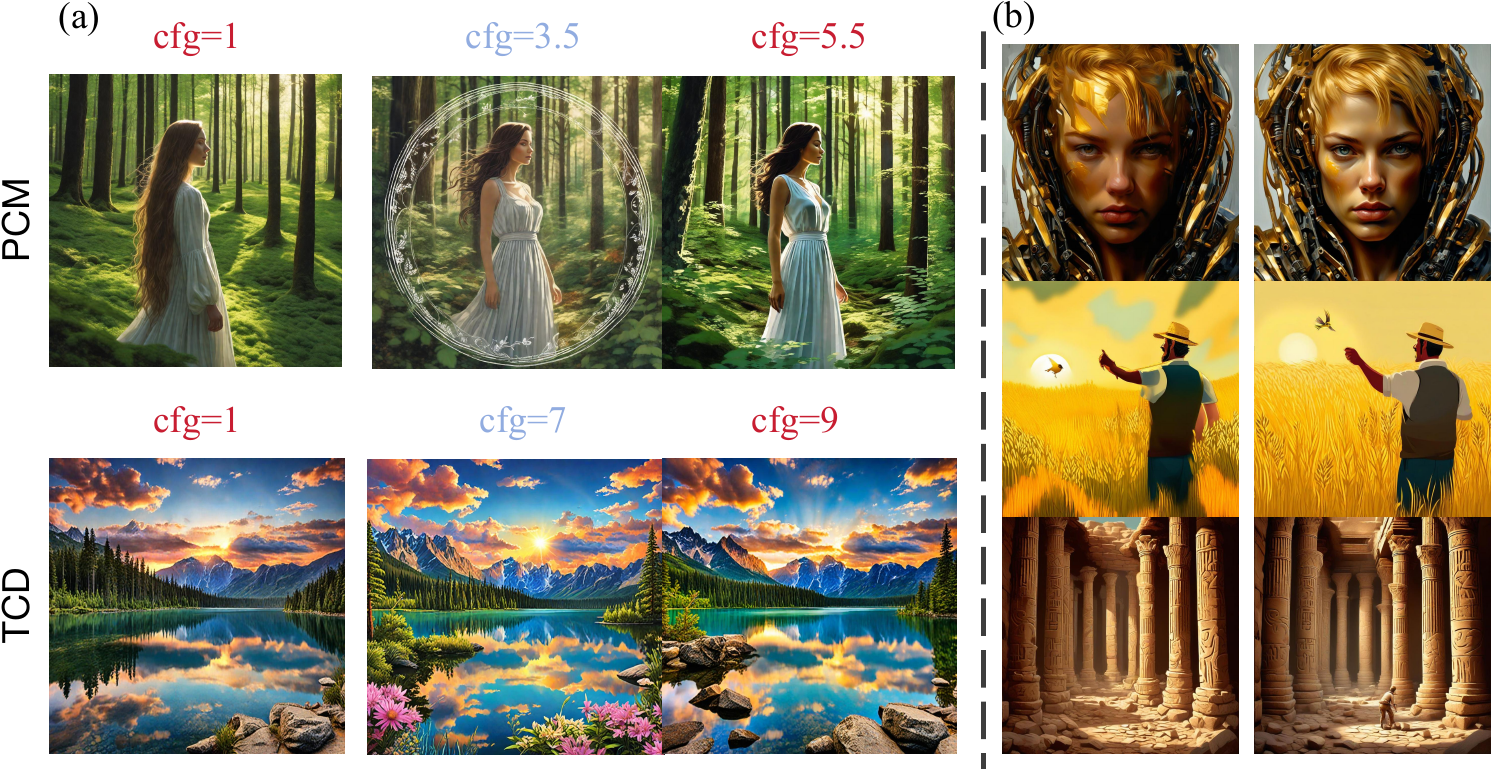}
    \caption{(a) Ablation comparison of distillation with decoupled guidance. (b) Ablation comparison of guidance scale tuning.}
    \label{fig:abs}
    \vspace{-10pt}
\end{figure}

To demonstrate the advantages of distillation with decoupled guidance, we conducted experiments with a batch size of 128, \( \mathcal{K}_{\min} = 4 \), \( \mathcal{K}_{\max} = 8 \), and \(\eta = 0.3\), training two models with empty prompt ratios of 0 and 0.2. After 15k steps, we performed inference using a CFG of 3, as shown in \Cref{fig:abs} (b). This approach effectively stabilized image quality and reduced visual artifacts. Additionally, we applied the guidance scale tuning to models like TCD and PCM, which were distilled with higher CFG values (corresponding to original CFGs of 9 and 5.5, respectively, when inferred with CFG 1). Guidance scale tuning successfully converted these models to use normal CFG values during inference, significantly enhancing image content richness by reducing the CFG.

\subsubsection{Effect of $\mathbf{x}_0$ Clipping Sample}
In Figure 6, we demonstrate the advantages of \(\mathbf{x}_0\) clipping. For some samples inferred with a low guidance scale, such as the one on the far left, certain defects may appear during inference. Increasing the guidance can alleviate these issues to some extent, but it also increases contrast, as seen in the middle image of Figure 6. By applying clipping in the initial steps, we can partially correct these defects without increasing contrast. Additional examples and results from applying clipping beyond the first denoising step are provided in the Appendix.
\section{Conclusion}

Consistency distillation methods have proven effective in accelerating diffusion models' generative tasks. However, previous methods often face issues such as blurriness and detail loss due to simplistic strategies in target timestep selection. We propose Target-Driven Distillation (TDD), which addresses these limitations by (1) employing a refined strategy for selecting target timesteps, thus enhancing training efficiency; (2) using decoupled guidance during training, which allows for post-tuning of the guidance scale during inference; and (3) incorporating optional non-equidistant sampling and $\mathbf{x}_0$ clipping for more flexible and precise image sampling. Experiments demonstrate that TDD achieves state-of-the-art performance in few-step generation, providing a superior option among consistency distillation methods.
\begin{figure}[t]
\centering
\includegraphics[width=1.\columnwidth]{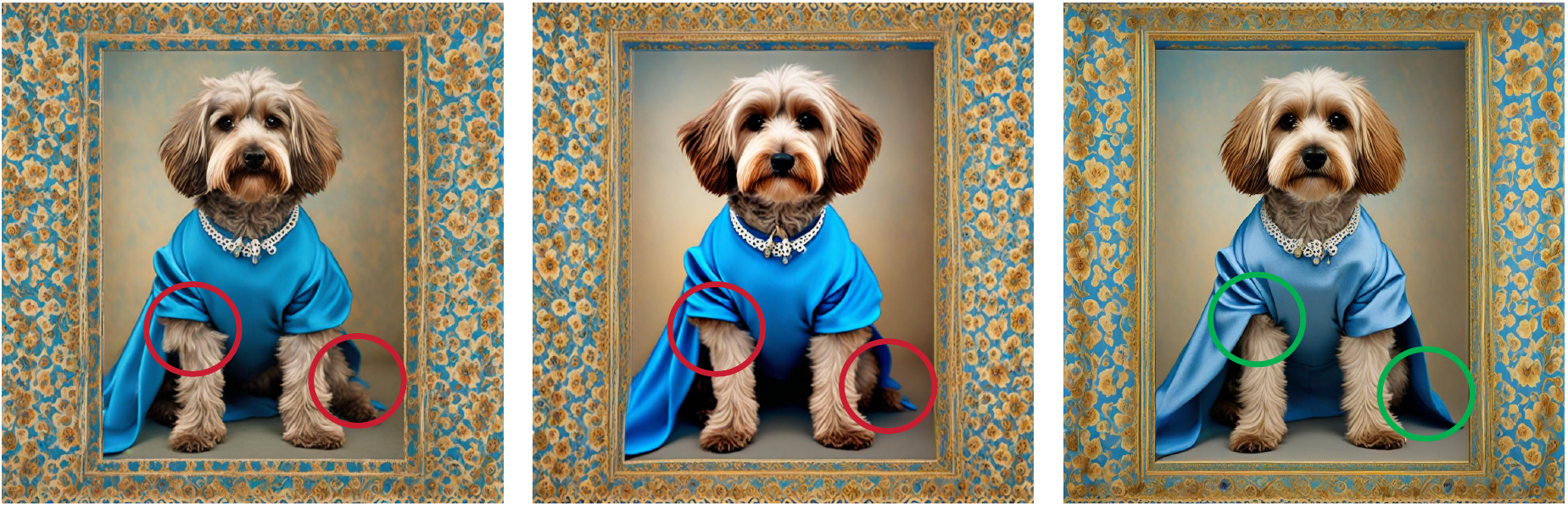}
\caption{Ablation comparison of $\mathbf{x}_0 $ clipping sample. The prompt used is ``a dog wearing a blue dress". The images on the left are with CFG = 2, the middle with CFG = 3, and the right with CFG = 3 and $\mathbf{x}_0 $ clipping applied.}
\label{fig_x0_clip}
\vspace{-10pt}
\end{figure}

\bibliography{aaai25}
\newpage
\begin{appendices}
\section{Appendix}
\subsection{Guidance Scale Tuning Details}
% \begin{}

Define $\epsilon_\theta(\mathbf{x}_t)$ as the noise predicted by the consistency model at each inference step. According to the Classifier-Free Guidance (CFG), the noise can be expressed as:
\begin{equation}
  \hat{\epsilon_w} = (1 + {w})\epsilon_\theta({\mathbf{x}}^{w'}_{t}, t, c) - {w}\epsilon_\theta({\mathbf{x}}^{w'}_{t}, t),  
\end{equation}
where $w'$ denotes the guidance scale used during the distillation process, and ${\mathbf{x}}^{w'}_{t}$ represents the sample at time step $t$ obtained from the model distilled with this guidance scale.

By replacing a portion of the conditions with empty prompts, we obtain the approximations:
\begin{equation}
\epsilon_\theta({\mathbf{x}}^{w'}_{t}, t, c) \approx (1 + {w'})\epsilon_\phi({\mathbf{x}}_{t}, t, c) - {w'}\epsilon_\phi({\mathbf{x}}_{t}, t) 
\end{equation} and
\begin{equation}
\epsilon_\theta({\mathbf{x}}^{w'}_{t}, t) \approx \epsilon_\phi({\mathbf{x}}_{t}, t),
\end{equation}
where $\epsilon_\phi(*)$ is the master model.Thus, we derive:
\begin{equation}
\begin{aligned}
\hat{\epsilon_w} \approx & (1 + {w})[ (1 + {w'})\epsilon_\phi({\mathbf{x}}_{t}, t, c) - {w'}\epsilon_\phi({\mathbf{x}}_{t}, t) 
] \\ &
- {w}\epsilon_\phi({\mathbf{x}}_{t}, t)
\\
\approx & (1 + w)(1 + {w'})\epsilon_\phi({\mathbf{x}}_{t}, t, c) 
 - (1 + w){w'}\epsilon_\phi({\mathbf{x}}_{t}, t)  \\ &
- {w}\epsilon_\phi({\mathbf{x}}_{t}, t) 
\\
\approx & (1 + {w'})[(1 + w)\epsilon_\phi({\mathbf{x}}_{t}, t, c) - w\epsilon_\phi({\mathbf{x}}_{t}, t)]
\\ & + (1 + {w'})w\epsilon_\phi({\mathbf{x}}_{t}, t) - (1 + w){w'}\epsilon_\phi({\mathbf{x}}_{t}, t) 
\\ & - {w}\epsilon_\phi({\mathbf{x}}_{t}, t)
\\
\approx & (1 + {w'})[(1 + w)\epsilon_\phi({\mathbf{x}}_{t}, t, c) - w\epsilon_\phi({\mathbf{x}}_{t}, t)]
\\ &  + w\epsilon_\phi({\mathbf{x}}_{t}, t) + {w'}w\epsilon_\phi({\mathbf{x}}_{t}, t) - {w'}\epsilon_\phi({\mathbf{x}}_{t}, t) 
\\ & - {w'}w\epsilon_\phi({\mathbf{x}}_{t}, t) 
- {w}\epsilon_\phi({\mathbf{x}}_{t}, t)
\\
\approx & (1 + {w'})[(1 + w)\epsilon_\phi({\mathbf{x}}_{t}, t, c) - w\epsilon_\phi({\mathbf{x}}_{t}, t)]
\\ & - {w'}\epsilon_\phi({\mathbf{x}}_{t}, t).
\end{aligned}
\end{equation}
Let \((1+w)\epsilon_\phi({\mathbf{x}}_{t}, t, c)-w\epsilon_\phi({\mathbf{x}}_{t}, t)\) denote $\epsilon_{w}$, which represents the noise predicted by the original model using the normal guidance scale $ w $ at each time step. We then obtain:
\begin{equation}
    \hat{\epsilon_w} \approx (1 + {w'})\epsilon_{w} - {w'}\epsilon_\phi({\mathbf{x}}_{t}, t).
\end{equation}
Given $\epsilon_{w}$ as the expected output, we get:
\begin{equation}
    \epsilon_{w} \approx [\hat{\epsilon_w} + {w'}\epsilon_\theta({\mathbf{x}}_{t}, t)]/(1+{w'}).
\end{equation}
Since ${w'}$ is a fixed constant and both $\epsilon_{w}$ and $\hat{\epsilon_w}$ are governed by the same guidance scale $w$, the formula allows us to tune the guidance scale to its normal range by incorporating an additional unconditional noise.

Additionally, for models distilled using a range of guidance scales, we can use the average of this range $\overline{w'}= {w'_{\min} + w'_{\max})/2}$, as a substitute for $w'$. The expression then becomes:

\begin{equation}
    \epsilon'_{w} \approx [\hat{\epsilon'_w} + \overline{w'}\epsilon'_\theta({\mathbf{x}}_{t}, t)]/(1+\overline{w'}).
\end{equation}
\begin{figure}[t]
    \centering
    \includegraphics[width=1.\linewidth]{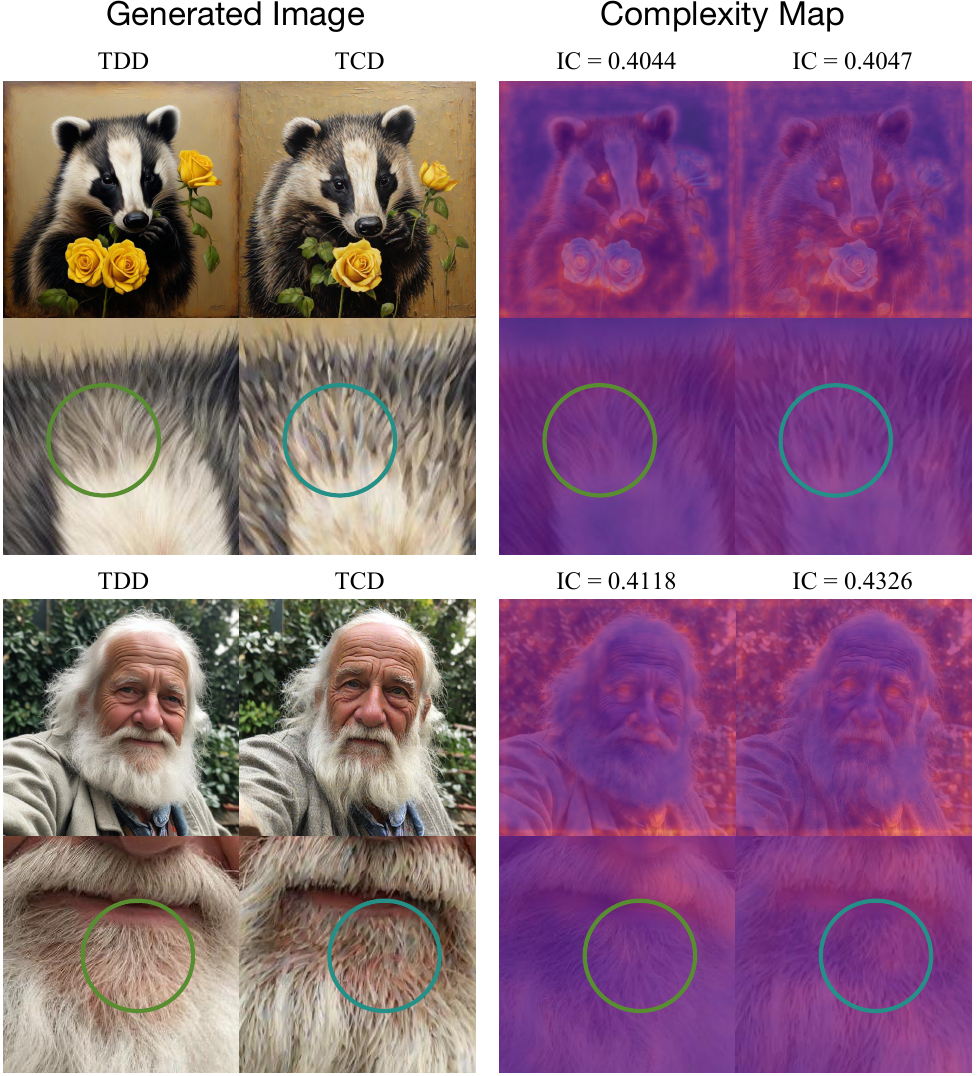}
    \caption{Comparison of image complexity. We present complexity maps for selected generated images from TCD and TDD.}
    \vspace{-10pt}
    \label{fig:ic_1}
\end{figure}
\subsection{Experimental Details}
\subsubsection{Datasets}
For our experiments, we utilize a carefully curated subset of the Laion-5B High-Resolution dataset, specifically selecting images with an aesthetic score exceeding 5.5. This subset comprises approximately 260 million high-quality images, providing a diverse and extensive foundation for training our models.
To evaluate our models' performance comprehensively, we employ the COCO-2014 validation set. This dataset is divided into two subsets: COCO-30K, containing 30,000 captions, and COCO-2K, with 2000 captions. These subsets are used to assess a range of metrics, ensuring a robust evaluation across different aspects of image captioning and understanding.
Additionally, we benchmark our models using the PartiPrompts dataset, which consists of over 1600 prompts spanning various categories and challenging aspects. This dataset is particularly valuable for testing the model's generalization and adaptability across diverse and complex scenarios.

\subsubsection{Training Details}
In our experiments, for the main results comparison, we utilized the SDXL LoRA versions of LCM, TCD, and PCM with open-source weights, where PCM was distilled with small CFG across 4 phases. For our model, we similarly chose SDXL as the backbone for distillation, setting the LoRA rank to 64. For the non-adversarial version, we employed a learning rate of 1e-6 with a batch size of 512 for 20,000 iterations. For the adversarial version, the learning rate was 2e-6 with the adversarial model set to 1e-5 and a batch size of 448.  $\mathcal{K}_{\min}$ and $\mathcal{K}_{\max}$ are set to 4 and 8, respectively, with $\eta$ set to 0.3 and the ratio of empty prompts set to 0.2.We utilized DDIM as the solver with $N=250$. Additionally, we used a fixed guidance scale of $w'=3.5$.

In our non-equidistant sampling method, we incrementally inserted timesteps from an 8-step denoising process into the intervals of a 4-step denoising process. Specifically, for timesteps between steps 4 and 8, the selection was as follows:
\begin{itemize}
    \item For 5 steps: [999., 875., 751., 499., 251.].
    \item For 6 steps: [999., 875., 751., 627., 499., 251.].
    \item For 7 steps: [999., 875., 751., 627., 499., 375., 251.].
\end{itemize}
For the ablation experiments, we consistently employed non-adversarial distillation approach with the following settings: a learning rate of 5e-6, a batch size of 128, and the DDIM solver with $N=50$. We trained all the ablation models using 15,000 iterations.

To ensure fairness in the main results comparison, given the varying CFG values used in prior work for distillation, we standardize the guidance scales for inference based on the distillation guidance scales used in LCM and TCD. Specifically, we use CFG = 1.0 for LCM and TCD, CFG = 1.6 for PCM, and CFG = 2.0 for our method.
\begin{figure}[t]
    \centering
    \includegraphics[width=1.\linewidth]{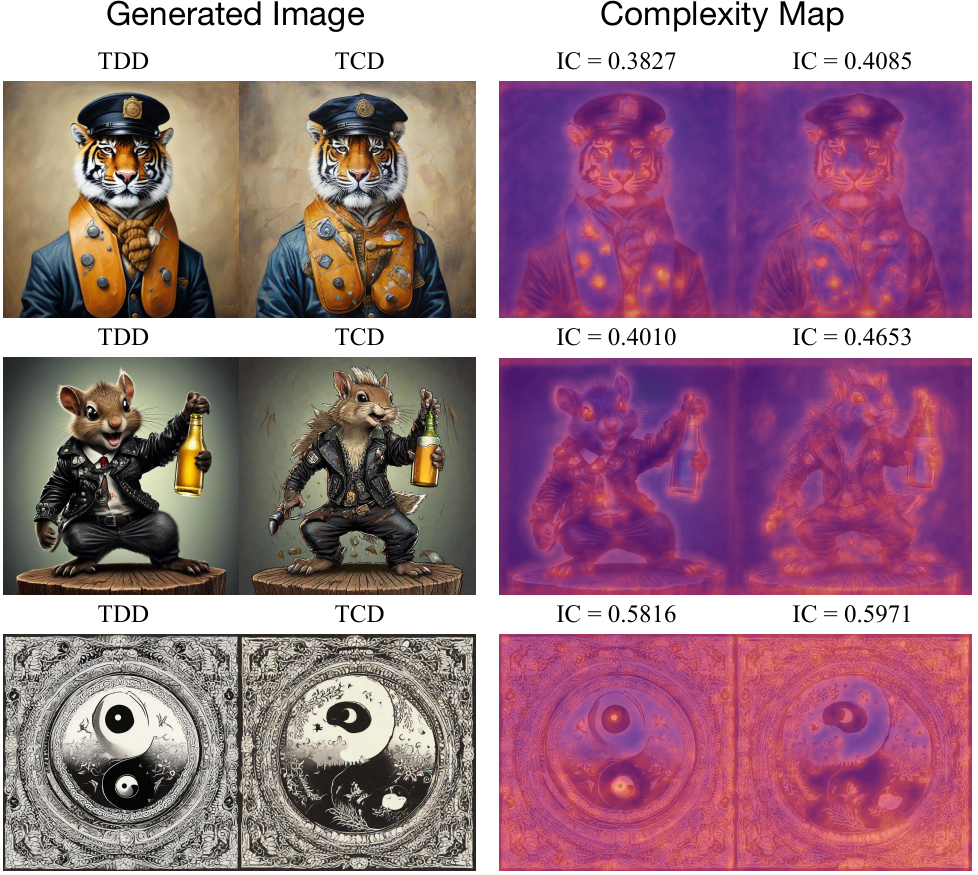}
    \caption{Further comparison of image complexity results.}
    \label{fig:ic_2}
    \vspace{-10pt}
\end{figure}
\subsubsection{Analyzing Image Complexity}

Although our model does not achieve the highest image complexity metrics, we identify factors that may influence this assessment. These factors primarily fall into two categories: visual artifacts and high-frequency noise, which can cause the IC model to misinterpret additional content, and unstable generation, which results in chaotic images that inflate the IC score. 

As shown in \Cref{fig:ic_1}, visual artifacts appear in animal fur and elderly facial hair, leading the model to mistakenly perceive these areas as more complex. This is merely a result of generation defects.
Additionally, as shown in \Cref{fig:ic_2}, the bodies of the tiger and mouse exhibit line irregularities and content disarray due to generation instability, which also contributes to inflated complexity metrics. The presence of extraneous lines and colors in the backgrounds further increases complexity. This issue is particularly pronounced in the ``Yin-Yang" image, where instability causes a more noticeable rise in complexity. However, this increased complexity is a result of defects rather than meaningful content. Our goal is to achieve clean, coherent, and meaningful image content. Thus, despite a slight decrease in image complexity, our method effectively balances image quality and content richness.
\begin{figure}[t]
    \centering
    \includegraphics[width=1.\linewidth]{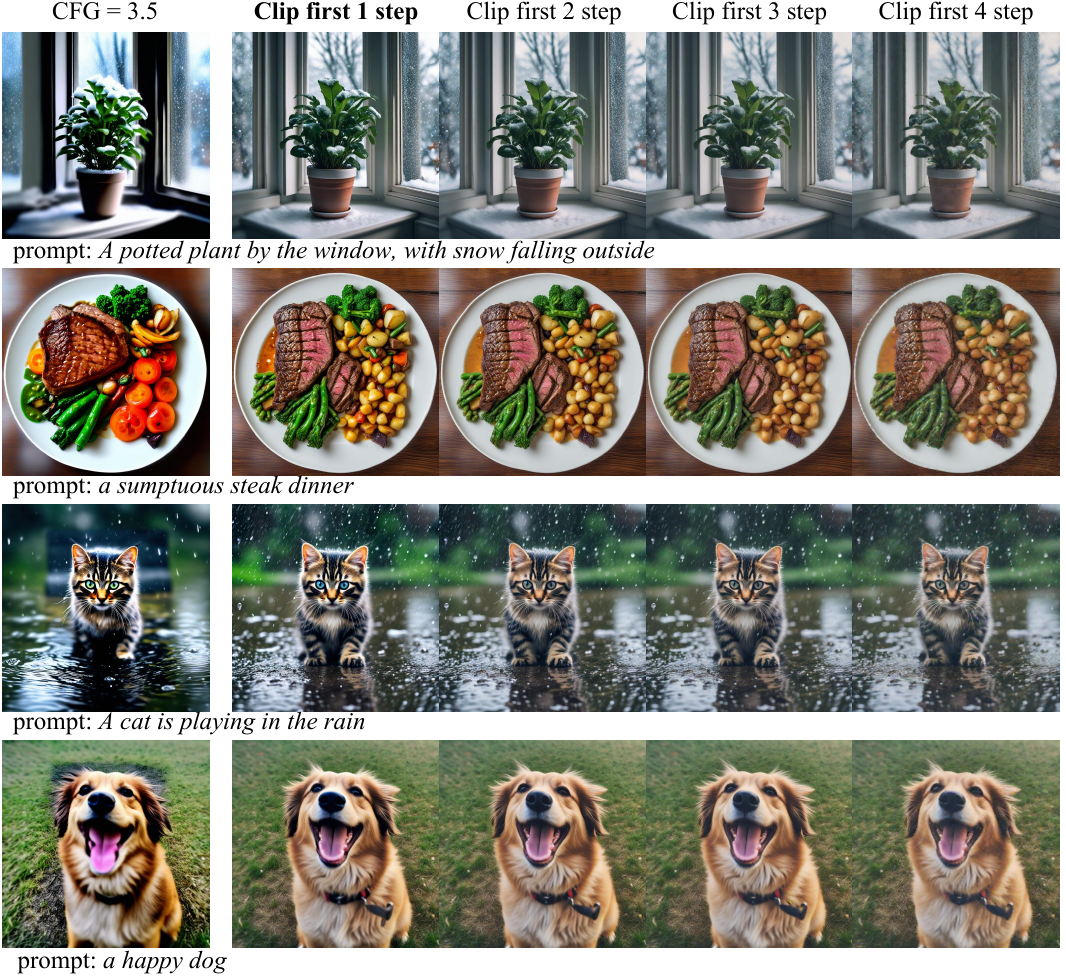}
    \caption{Qualitative results of TDD with 4-step inference, comparing the effect of applying $\mathbf{x}_0$ clipping for different numbers of initial steps.}
    \vspace{-10pt}
    \label{fig:x0}
\end{figure}
\subsubsection{$\mathbf{x}_0$ Clipping Sample Details}

We further examine the $\mathbf{x}_0$ clipping sample using the TDD model (non-adversarial) with a guidance scale of 3.5 and 4-step sampling. Without $\mathbf{x}_0$ clipping, the image exhibits excessively high contrast, resulting in an overall unrealistic appearance. In contrast, when applying $\mathbf{x}_0$ clipping at the initial step, the image becomes more natural and reveals additional details, such as the scenery outside the window and the garnishes in the food, as shown in \Cref{fig:x0}. However, increasing the number of steps where $\mathbf{x}_0$ clipping is applied, from just the first step to every step, results in images that progressively become more washed out and blurry. Subsequent $\mathbf{x}_0$ clipping operations do not enhance realism but instead lead to reduced clarity. Therefore, we recommend applying $\mathbf{x}_0$ clipping only at the first step for higher CFG values to ensure an improvement in image quality.

\subsubsection{More Visualizations}
\begin{figure*}
    \centering
    \includegraphics[height=1.\textheight]{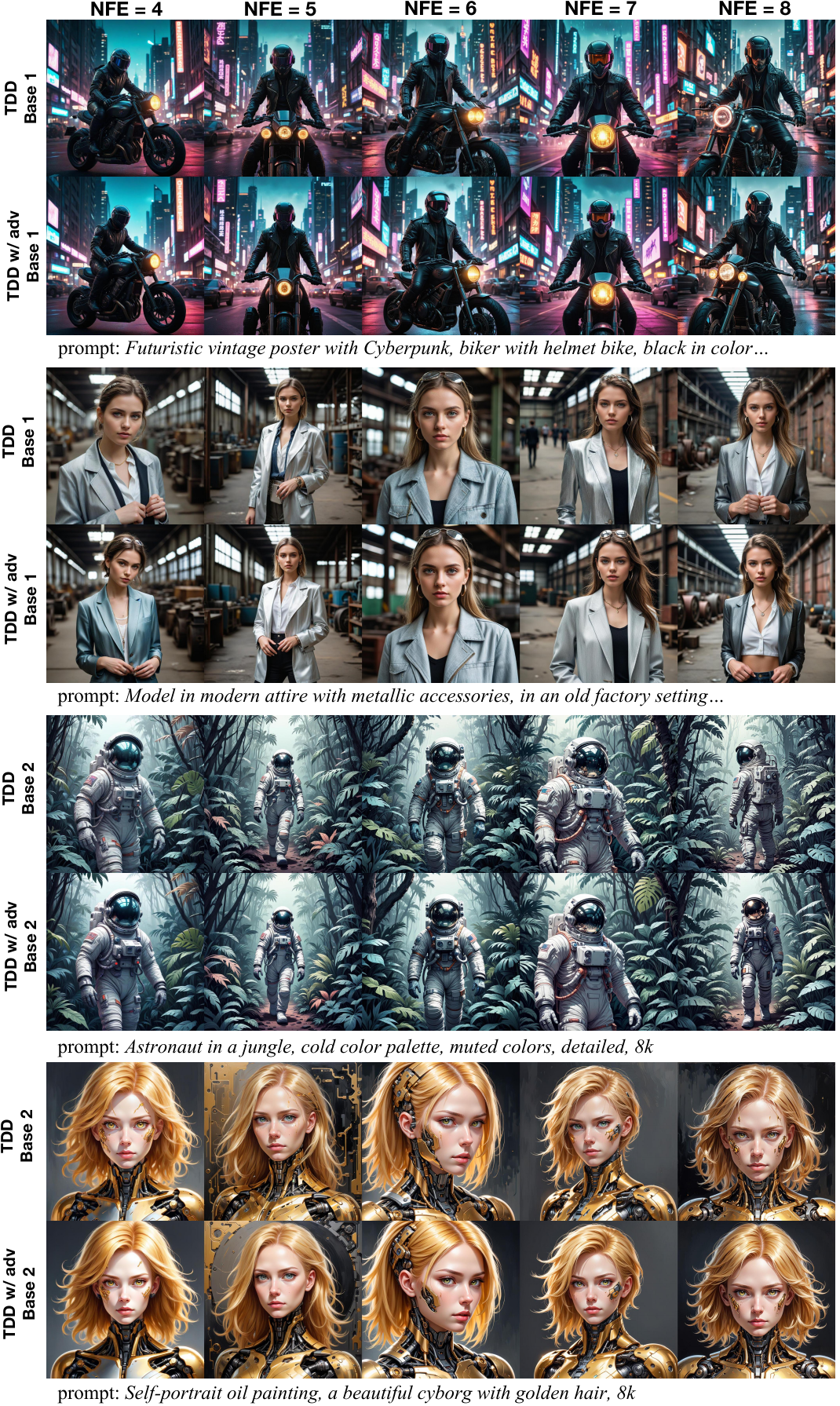} % 替换为你的图像文件名
    \caption{Qualitative results of TDD using different base models. Base 1: realvisxlV40; Base 2: SDXLUnstableDiffusers-YamerMIX.}
    \label{fig:diff_base}
\end{figure*}

\begin{figure*}
    \centering
    \includegraphics[height=1.\textheight]{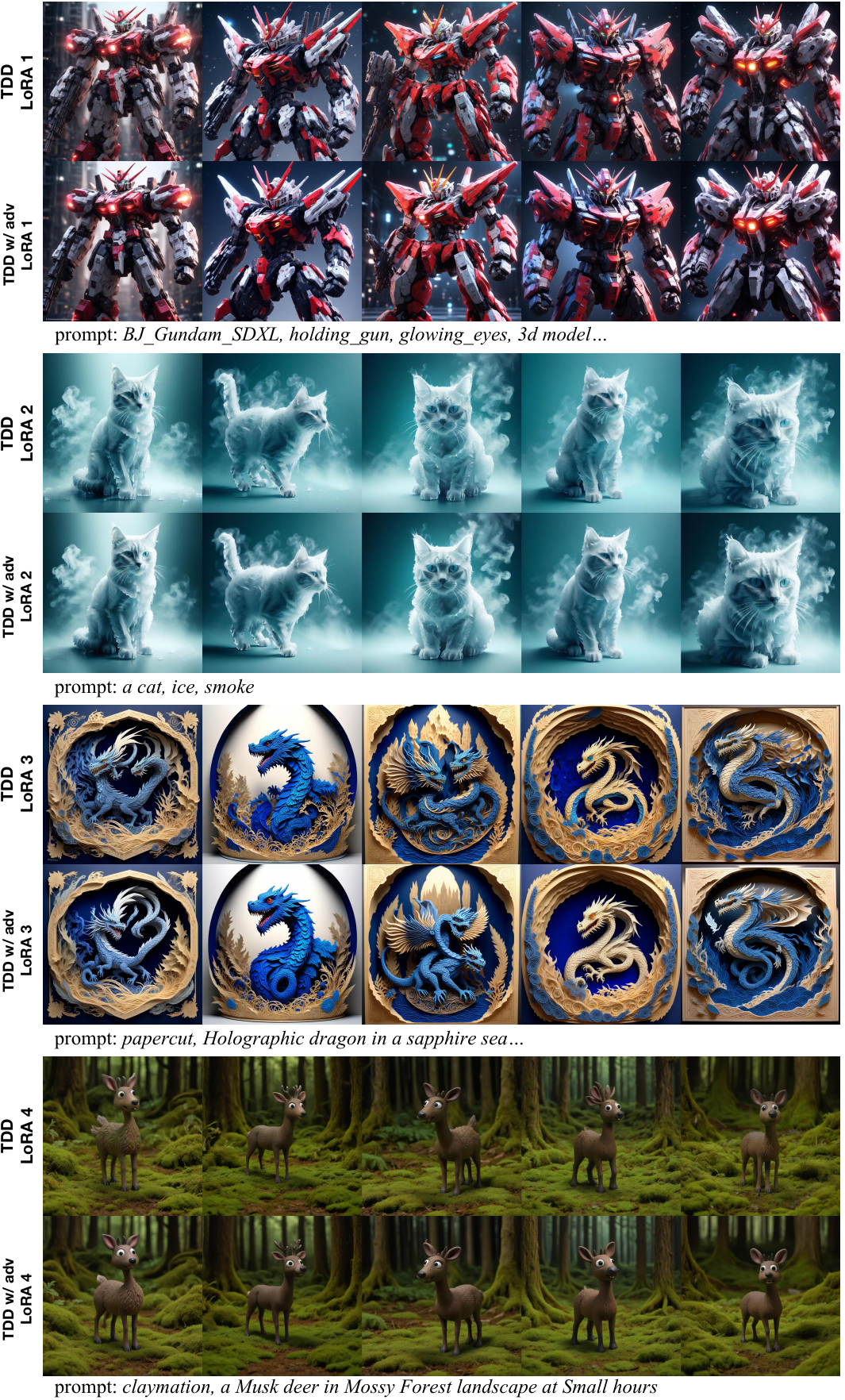} % 替换为你的图像文件名
    \caption{Qualitative results of TDD using different LoRA adapters with 4-step inference. LoRA 1: SDXL-GundamV3; LoRA 2: Ice; LoRA 3: Papercut; LoRA 4: CLAYMATEV2.03.}
    \label{fig:diff_lora}
\end{figure*}

\begin{figure*}
    \centering
   \includegraphics[height=1.\textheight]{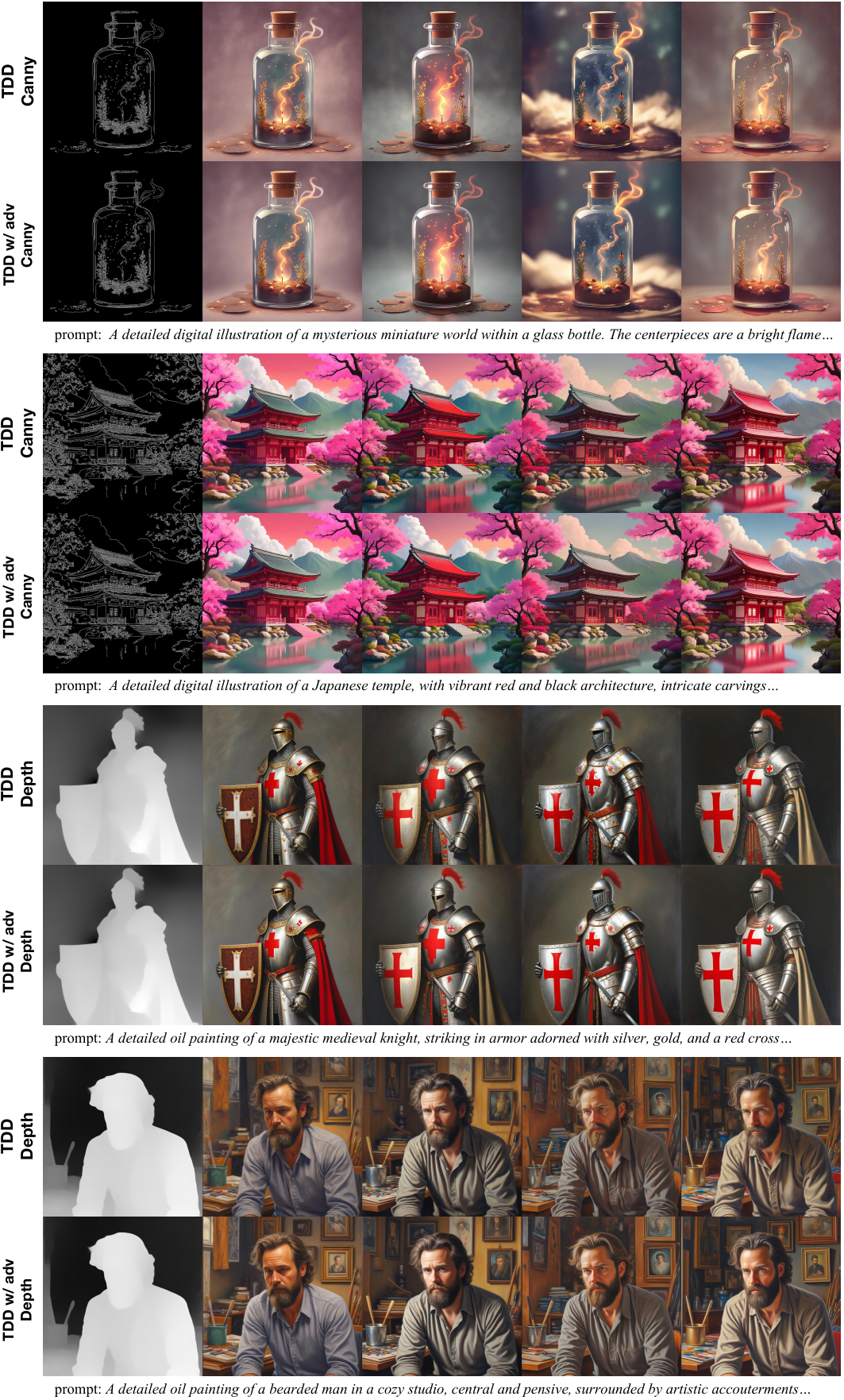} % 替换为你的图像文件名
    \caption{Qualitative results of TDD using ControlNets based on Canny (top) and Depth (bottom) with 4-step inference.}
    \label{fig:diff_ctrl}
\end{figure*}

\begin{figure*}
    \centering
    \includegraphics[height=1.\textheight]{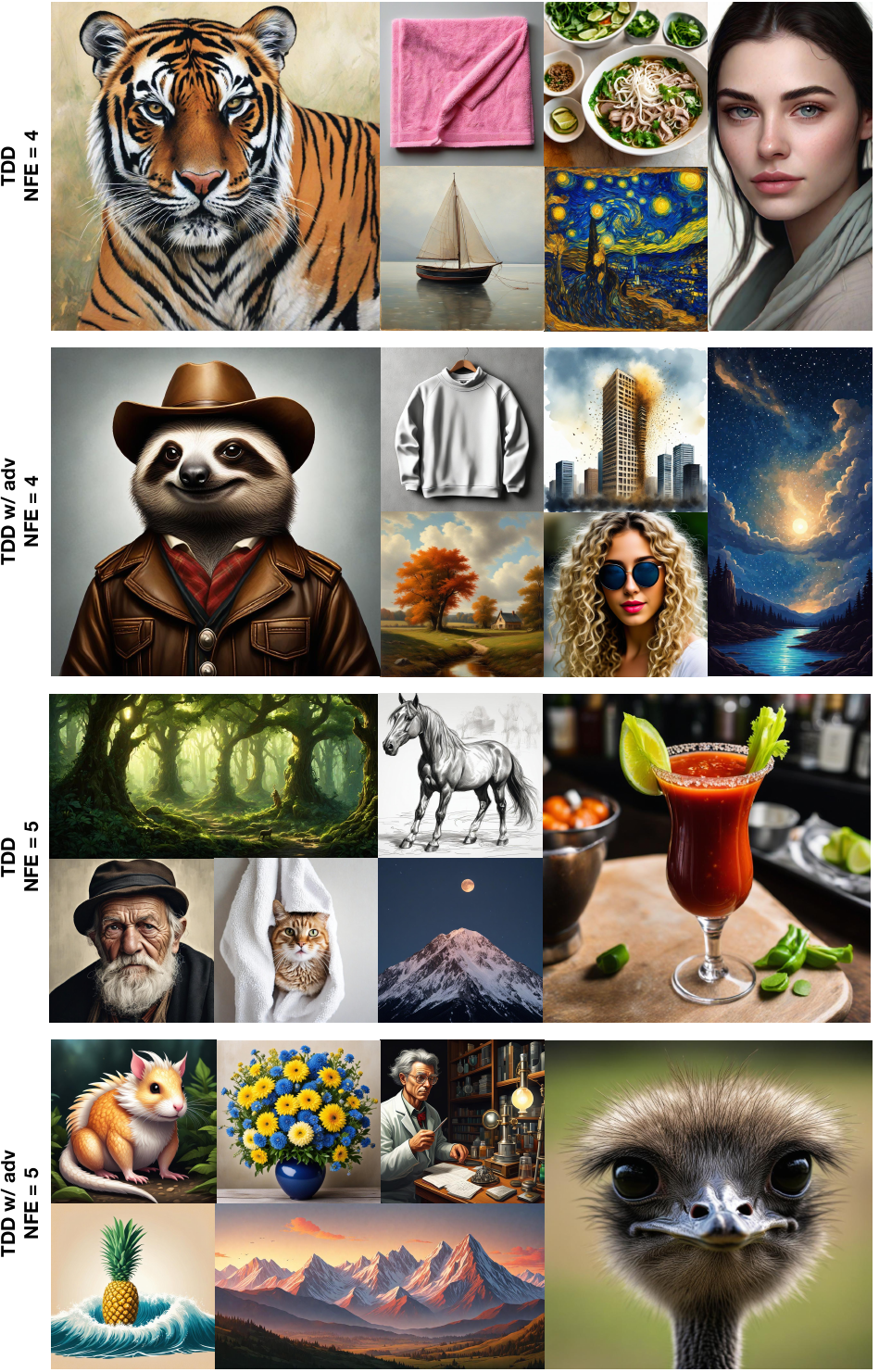} % 替换为你的图像文件名
    \caption{Samples generated by TDD using four or five steps with Stable Diffusion XL.}
    \label{fig:step4_5}
\end{figure*}

\begin{figure*}
    \centering
    \includegraphics[height=1.\textheight]{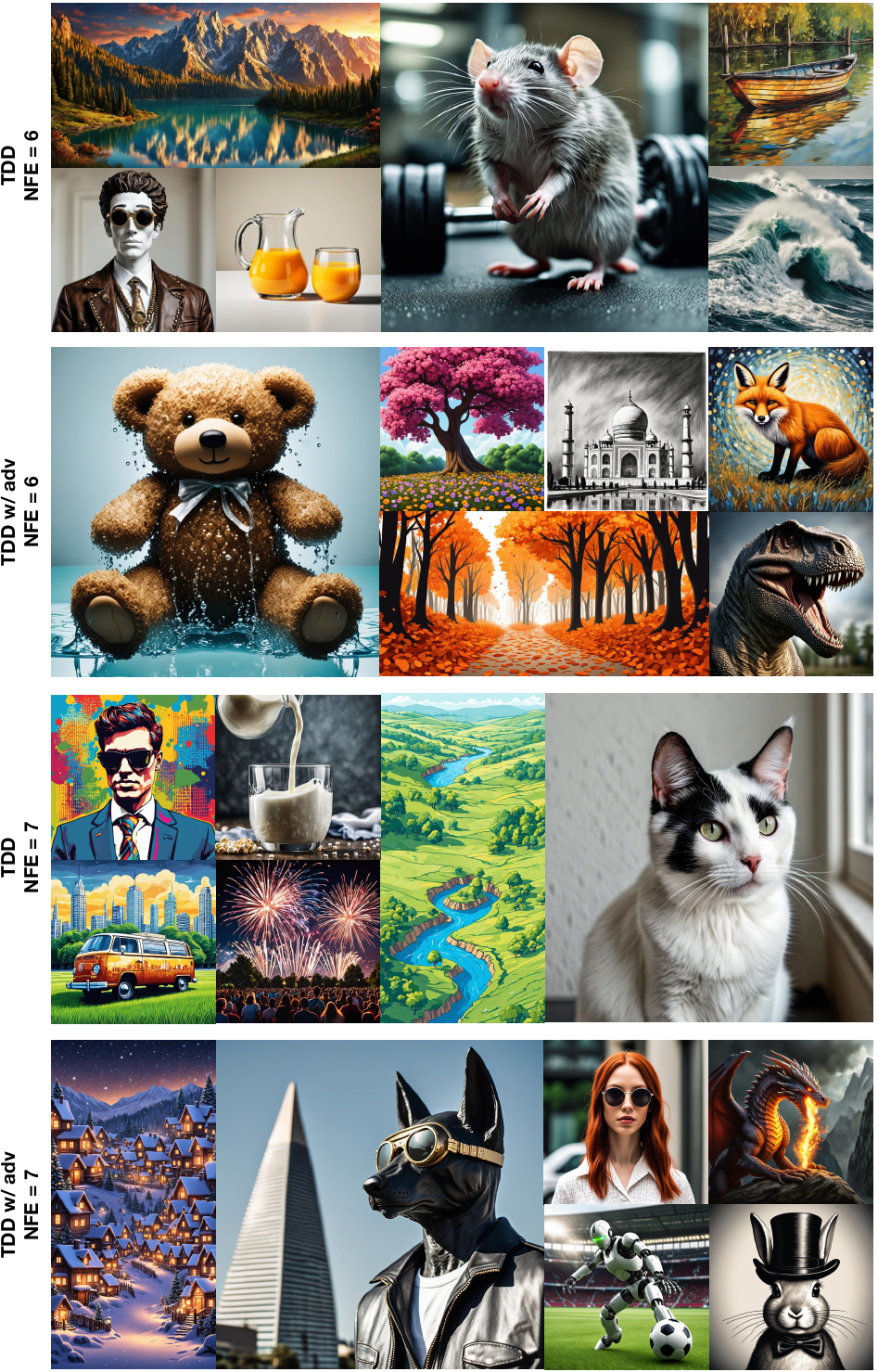} % 替换为你的图像文件名
    \caption{Samples generated by TDD using six or seven steps with Stable Diffusion XL.}
    \label{fig:step6_7}
\end{figure*}

\begin{figure*}
    \centering
    \includegraphics[width=1.\linewidth]{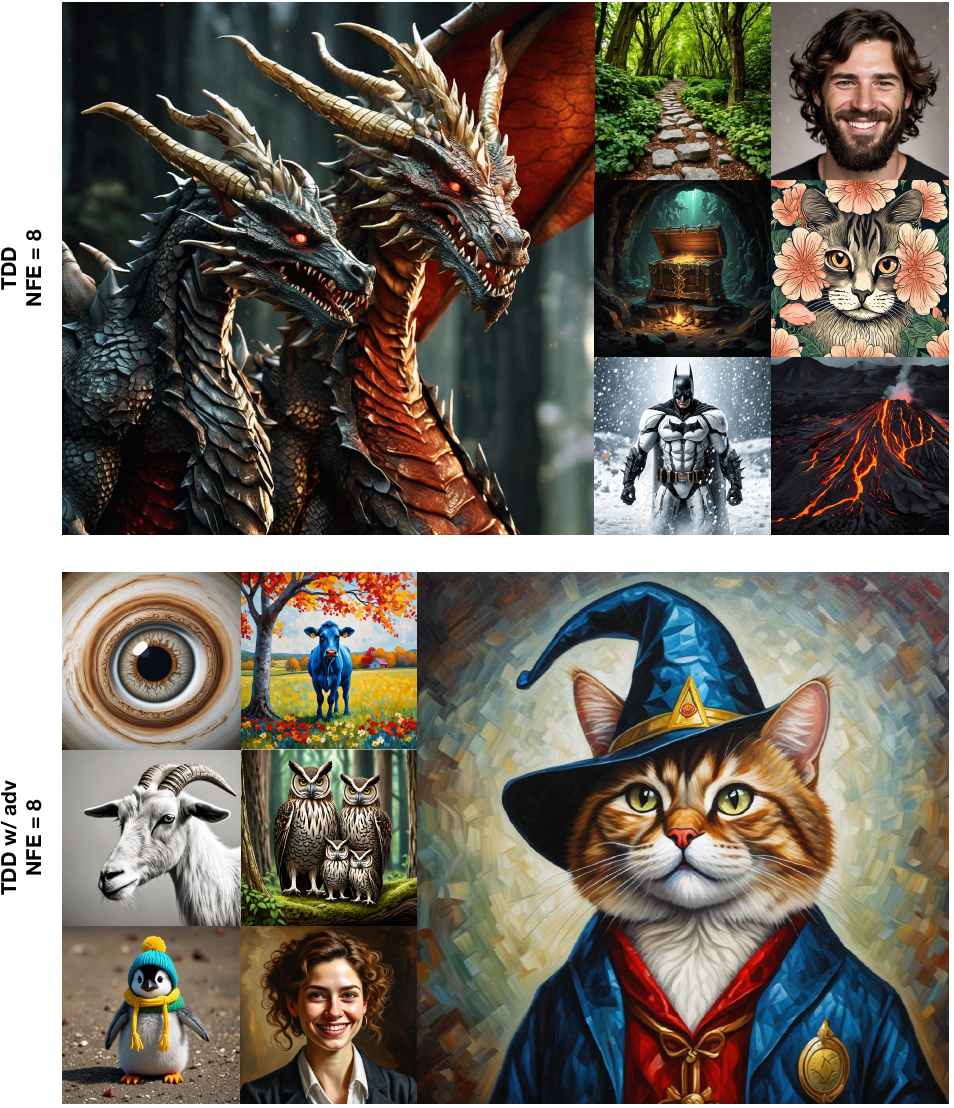} % 替换为你的图像文件名
    \caption{Samples generated by TDD using eight steps with Stable Diffusion XL.}
    \label{fig:step8}
\end{figure*}

We also illustrate some additional samples:
\begin{itemize}
    \item using other base models finetuned from SDXL in \Cref{fig:diff_base};
    \item using LoRA adapters in \Cref{fig:diff_lora};
    \item controlled by ControlNet in \Cref{fig:diff_ctrl};
    \item with different number of steps in \Cref{fig:step4_5} to \Cref{fig:step8}.
\end{itemize}

\clearpage
\clearpage

\clearpage

\clearpage

\clearpage

\clearpage

\end{appendices}

\end{document}